\documentclass[final,5p,times,twocolumn]{elsarticle}

\biboptions{numbers,sort&compress}

\usepackage{amssymb}
\usepackage{amsmath}

\usepackage{lineno}
\usepackage{rotating}
\usepackage{booktabs}
\usepackage{caption}
\usepackage{soul}
\usepackage{xcolor}
\usepackage[table]{xcolor}
\sethlcolor{yellow!70}
\usepackage{epsfig}
\usepackage{multirow}
\usepackage{hyperref}
\usepackage{orcidlink}

\journal{Journal of Manufacturing Systems}

\begin{document}

\begin{frontmatter}

\title{Kolmogorov-Arnold Networks-Based Tolerance-Aware Manufacturability Assessment Integrating Design-for-Manufacturing Principles}

\author{Masoud Deylami\,\orcidlink{0009-0006-7136-4583}}

\author{Negar Izadipour\,\orcidlink{0009-0003-6261-0031}}

\author{Adel Alaeddini\,\orcidlink{0000-0003-4451-3150}\corref{cor1}}
\ead{aalaeddini@mail.smu.edu}

\cortext[cor1]{Corresponding author.}

\affiliation{organization={Department of Mechanical Engineering, Southern Methodist University},
            addressline={3101 Dyer St.}, 
            city={Dallas},
            postcode={75205}, 
            state={TX},
            country={USA}}

\begin{abstract}
\label{sec1_abstract}

Manufacturability assessment is a critical step in bridging the persistent gap between design and production. While artificial intelligence (AI) has been widely applied to this task, most existing frameworks depend on geometry-driven methods that require extensive preprocessing, suffer from information loss, and provide limited interpretability. This study proposes a new methodology that evaluates manufacturability directly from parametric design features, enabling the explicit incorporation of dimensional tolerances into the analysis. The approach employs Kolmogorov–Arnold Networks (KANs) to learn functional relationships between design parameters, tolerances, and manufacturability outcomes without requiring computer-aided design (CAD) processing. A synthetic dataset of 300,000 labeled designs was generated to demonstrate the method performance across three representative scenarios, hole drilling, pocket milling, and combined drilling–milling, while accounting for machining constraints and design-for-manufacturing \allowbreak (DFM) rules. Benchmarking against fourteen machine learning (ML) and deep learning (DL) models shows that KAN achieves the highest performance in all scenarios, with AUC values of 0.9919 for drilling, 0.9841 for milling, and 0.9406 for the combined case. The proposed approach offers high interpretability through spline-based functional visualizations and KAN-derived latent-space projections, enabling precise identification of the geometric and tolerance parameters that most strongly influence manufacturability. An industrial case study further demonstrates how the proposed framework enables iterative, parameter-level design modifications that successfully transform a non-manufacturable component into a manufacturable one, with model-predicted defects corroborated through physical fabrication. The results demonstrate that the proposed methodology provides a lightweight, transparent, scalable, and computationally efficient alternative to geometry-driven frameworks, delivering quick and actionable manufacturability feedback fully aligned with engineering design principles and semantics.

\end{abstract}


\end{frontmatter}


\section{Introduction}
\label{sec2_introduction}

Manufacturability assessment is a key link between design and manufacturing~\cite{yan2024, chen2025manu}, determining whether a given design can be produced effectively and economically. However, novice designers and those without manufacturing experience often struggle to evaluate manufacturability~\cite{doellken2020}. These difficulties stem from a lack of manufacturing knowledge and experience~\cite{jing2026, jarosz2024}, specific process constraints~\cite{abdelall2020}, challenges in developing cost-optimized and manufacturing-oriented designs~\cite{ordek2025}, and the inability to accurately evaluate manufacturing feasibility~\cite{zhang2022web}.

\begin{figure*}[t]
    \centering
    \includegraphics[width=1\linewidth]{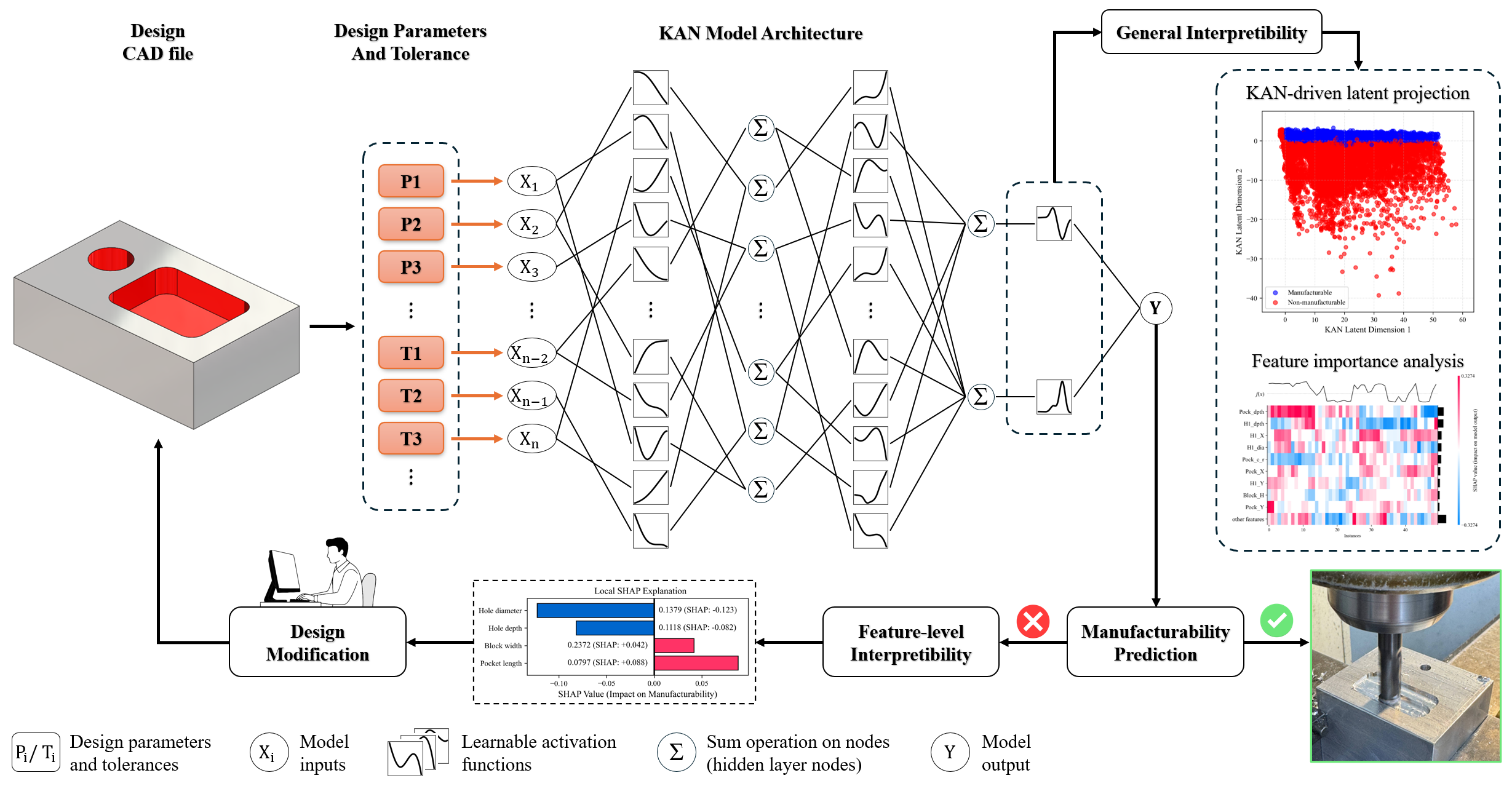}
    \caption{Proposed KAN-based manufacturability assessment and diagnosis methodology workflow.}
    \label{fig1_flowdiagram}
\end{figure*}

With the transition from Industry 4.0 to Industry 5.0~\cite{xu2021, leng2023}, design and manufacturing paradigms have evolved from automation-driven systems toward more human-centric frameworks~\cite{leng2022, zhang2023}. In this paradigm shift, AI plays a pivotal role. Within contemporary Industry 5.0 manufacturing ecosystems, AI enables advancements in quality control~\cite{chen2025}, defect and anomaly detection~\cite{diogo2025}, predictive maintenance and condition monitoring~\cite{yedurkar2025}, production planning and scheduling~\cite{hameed2023}, supply chain optimization~\cite{feng2025}, digital twin (DT) technology~\cite{zeynivand2025}, process parameter optimization~\cite{wan2025}, additive manufacturing (AM) optimization~\cite{luo2025}, and error and fault diagnosis~\cite{gultekin2025}, among others. Similarly, in modern design, particularly within CAD environments, AI supports generative design~\cite{tanveer2025, wang2024}, topology optimization~\cite{cai2025}, CAD classification and clustering~\cite{krahe2022similarity}, design optimization~\cite{wang2025}, 3D model reconstruction~\cite{zhang2025cad}, CAD retrieval~\cite{jones2023cad}, and evaluation of design alternatives~\cite{li2025genai}.

Within this evolving landscape, AI-based manufacturability assessment has become a crucial research topic that aims to fill the long-standing gap between design and manufacturing by augmenting human capabilities and transforming manufacturability assessment from a rigid, rule-based process into an intelligent, adaptive, and data-driven decision-support function~\cite{xu2024, picard2025}. Most studies exploring AI-driven frameworks for manufacturability assessment utilize convolutional neural networks (CNNs) and rely on processing CAD files, utilizing geometric deep learning (GDL) techniques~\cite{heidari2025gdl}. Although GDL methods can process inputs in mesh~\cite{feng2019meshnet}, voxel~\cite{zhang2018featurenet}, and point cloud~\cite{le2018pointgrid} data formats, directly processing files in boundary-representation (B-Rep) format remains challenging. This limitation requires converting B-Rep models into mesh representations, a process that is computationally expensive, prone to information loss~\cite{jayaraman2021uvnet}, and unable to accurately capture fine geometric micro-features such as fillets and chamfers within the CAD file.

To address these limitations, this study presents a new and highly interpretable approach for assessing manufacturability based on KANs~\cite{liu2024kan}, which uses a tabular representation of designs rather than geometric CAD files. \hyperref[fig1_flowdiagram]{\textcolor{blue}{Figure~\ref*{fig1_flowdiagram}}} demonstrates the proposed methodology, in which each design is represented by its parametric attributes, capturing both geometric and functional features without the need for CAD processing or discretization. This method prevents information loss due to discretization and format conversion, removes resolution dependence, and reduces data augmentation costs. It also significantly improves computational efficiency and enables the consideration of tolerances in the evaluation process. More importantly, it significantly improves interpretability revealing which design parameters most strongly influence manufacturability outcomes. By integrating predictive accuracy with explanatory capability, the proposed method connects complex black-box AI models with human-centered design feedback, advancing the goal of interpretable, data-efficient manufacturability evaluation in the Industry~5.0 era. The primary contributions of this research are summarized as follows:

\begin{enumerate}

    \item A novel KAN-based manufacturability evaluation approach is proposed, providing substantial improvements in computational efficiency and inference speed compared to geometry-driven GDL methods that rely on CAD processing.

    \item A large-scale, scenario-specific synthetic dataset was developed for three representative manufacturing operations: \textit{hole drilling}, \textit{pocket milling}, and \textit{combined drilling-milling}. This dataset was created using a Python-based parametric modeling script that incorporates machining constraints and DFM rules. The resulting dataset contains over 300,000 labeled design instances.

    \item A new parametric data structure for manufacturability analysis is introduced, offering high interpretability and direct alignment with engineering design semantics. This representation allows manufacturability assessments to be directly mapped back into CAD models, supporting seamless post-assessment design modification.
    
    \item The practical reliability of the proposed approach was experimentally validated through a design case study involving progressive evaluation and parameter modifications, where a representative component with complex non-manufacturability conditions was successfully analyzed over multiple iterations, refined through this iterative process, and ultimately manufactured.
    
    \item Dimensional tolerances were explicitly incorporated into the dataset across all relevant geometric parameters, capturing their critical influence on manufacturability and enabling a more realistic representation of industrial design conditions.

    \item Enhanced interpretability is achieved through spline-function analysis and KAN-derived latent-space projections, enabling precise identification of the geometric and tolerance parameters that most strongly contribute to non-manufacturability and delivering actionable guidance for targeted design refinement.

\end{enumerate}

The remainder of this paper is organized as follows. \hyperref[sec3_related_works]{\textcolor{blue}{Section~\ref*{sec3_related_works}}} provides a comprehensive literature review of the relevant studies. \hyperref[sec4_data_generation]{\textcolor{blue}{Section~\ref*{sec4_data_generation}}} describes the proposed data structure and methodology for data generation and synthetic dataset creation. \hyperref[sec5_methodology]{\textcolor{blue}{Section~\ref*{sec5_methodology}}} presents the manufacturability assessment methodology, including a detailed description of the network structure and architecture, training procedure, and interpretability tools used in this study. \hyperref[sec6_results]{\textcolor{blue}{Section~\ref*{sec6_results}}} discusses the evaluation of the training process and the effectiveness of the proposed approach. \hyperref[sec7_CaseStudy]{\textcolor{blue}{Section~\ref*{sec7_CaseStudy}}} introduces the case studies as proof of concept. Finally, \hyperref[sec8_discussion]{\textcolor{blue}{Section~\ref*{sec8_discussion}}} concludes the paper and outlines directions for future research.

\section{Related Works}
\label{sec3_related_works}

\subsection{AI-based Manufacturability Assessment Approaches}
\label{sec3_1_relwork_AI}

Manufacturability evaluation of designs has long been a challenging topic\allowbreak ~\cite{madan2007}, as it bridges design and manufacturing criteria and must be addressed during the early design stages to avoid unnecessary waste of time, cost, and resources. With the advent of advanced computing technologies and the rise of AI, researchers have increasingly applied machine learning techniques across various engineering domains. The introduction of modern AI architectures, such as CNNs, has drawn significant attention to leveraging these high-potential models for specialized tasks, including the evaluation of manufacturability in CAD designs. While some studies have concentrated on subtractive manufacturing operations~\cite{yan2024}, others have focused on AM techniques~\cite{zhang2022}, broader manufacturing criteria~\cite{hwang2024, siegfried2024}, or the development of tools and software dedicated to manufacturability assessment~\cite{lynn2016, campana2020_1, winkler2021, xu2022, stavropoulos2022}.

Early attempts to employ neural networks (NNs) for manufacturability and design feasibility analysis involved the development of neuro-fuzzy frameworks. One of the earliest studies was conducted by Korosec~et~al.~\cite{korosec2005}, who utilized a feedforward neural network (FNN) to develop a neuro-fuzzy framework. They introduced a machining complexity index that integrates geometric parameters with technological attributes to evaluate the manufacturability of freeform surfaces.

In subtractive manufacturing, Ghadai~et~al.~\cite{ghadai2018_3DCNN} established a foundational approach for applying three-dimensional convolutional neural networks (3D-CNNs) directly to CAD geometry through voxelization of CAD files. The developed framework was capable of learning localized geometric features from voxelized CAD models augmented with surface normals. They also introduced a three-dimensional Gradient-weighted Class Activation Mapping (3D-GradCAM) technique to visualize non-manufacturable regions. Building upon this foundation, Peddireddy~et~al.~\cite{peddireddy2021_3DCNNs} extended 3D-CNN applications to multiple machining processes, proposing a two-step machining process identification (MPI) system using deep 3D-CNNs to automatically determine part manufacturability and classify the machining process type, milling or turning, directly from CAD models. Yan~et~al.~\cite{yan2022_3DVAEGANs} further advanced this field by introducing the 3D-VAE-GAN framework, which combines three-dimensional variational autoencoders (VAEs) and generative adversarial networks (GANs) to exploit shape transformation capability as a key factor in process identification and manufacturability assessment for lathe-based operations. Their study marked a shift from discriminative to generative modeling, representing machining process capabilities probabilistically rather than through predefined classification boundaries. In subsequent work, Yan~et~al.~\cite{yan2023_AESNN} proposed an autoencoder–Siamese neural network (AE-SNN) framework that integrates deep generative modeling with metric learning to automate manufacturability analysis and machining process selection for lathe- and mill-based components. This framework enabled quantitative manufacturability assessment through similarity-based deep learning, wherein design manufacturability is inferred by comparing query designs with learned process capabilities. Extending their generative modeling paradigm, Yan~et~al.~\cite{yan2024} later developed a DL-based semantic segmentation framework for distinguishing machinable and non-machinable regions and identifying candidate machining operations in 3D CAD models. More recently, Zhong~et~al.~\cite{zhong2025} introduced \textit{DeepMill}, a CNN-based neural accessibility learning framework designed for real-time cutter accessibility and manufacturability analysis in subtractive manufacturing.

In additive manufacturing, Shi~et~al.~\cite{shi2018} employed the heat kernel signature (HKS) method to develop a feature recognition (FR)–based manufacturability analysis framework capable of directly identifying geometric features and AM–specific constraints from 3D mesh models. Mycroft~et~al.~\cite{mycroft2020} introduced a ML–based framework for predicting printability in electron beam melting (EBM), with a focus on geometry-induced defects. Guo~et~al.\allowbreak  ~\cite{guo2021} integrated voxel-based geometric representations with stacked denoising autoencoders (AE) and an AE-GAN classifier to predict the manufacturability of metal cellular structures fabricated via direct metal laser sintering (DMLS) using limited labeled data. Zhang~et~al.~\cite{zhang2021} proposed a hybrid ML framework that combines a 3D-CNN for design analysis with an FNN for process parameter modeling to predict printability in laser powder bed fusion (LPBF). In subsequent studies, they combined CNNs with semantic segmentation to identify non-printable regions~\cite{zhang2022} and integrated 3D-CNNs with graph neural networks (GNNs) to assess the manufacturability of prismatic parts~\cite{zhang2022web}. Chen~et~al.~\cite{chen2025manu} later introduced a multilayer perceptron (MLP)–based self-supervised pretraining framework, termed volume-informed representation learning (VIRL), for few-shot manufacturability estimation, employing a graph-based encoder–decoder architecture with volumetric mapping.

Beyond additive and subtractive processes, several studies have explored forming-based manufacturing methods for both metallic and composite materials. Zimmerling~et~al.~\cite{zimmerling2019} developed a CNN–based framework for predicting textile drapability in continuous fiber-reinforced plastic (CoFRP) components by reproducing full-field forming behavior. Hwang~et~al.~\cite{hwang2024} proposed a ML–assisted formability prediction framework for the deep drawing of aluminum cups, integrating finite element (FE) simulations with ensemble learning classifiers. Siegfried~et~al.~\cite{siegfried2024} introduced a DL–based manufacturability analysis system for aluminum extrusion processes, combining a conditional multilayer perceptron (CMLP) and a CNN to predict extrusion pressure and perform profile similarity searches in support of feasibility analysis.

\subsection{Non–AI-based Manufacturability Evaluation Approaches}
\label{sec3_2_relwork_Non_AI}

Several studies have investigated the manufacturability of designs without employing AI-based technologies. Nelaturi~et~al.~\cite{nelaturi2015} introduced a printability map based on mathematical morphology and the medial axis transform to support manufacturability assessment and model correction feedback in AM. Cai~et~al.~\cite{cai2018} proposed a freeform machining FR framework that decomposes CAD models into discrete differential geometry elements and applies tool type, shape, and accessibility analyses to ensure manufacturability. Gupta~et~al.~\cite{gupta2019} developed a computer-aided manufacturability evaluation system for prismatic machining parts with orthogonal features, employing a FR method that extracts and classifies machining features from STEP-based CAD models. Chen~et~al.~\cite{chen2020} introduced a mathematical manufacturability metric that integrates machinability and printability to evaluate tool and nozzle accessibility. Coatanéa~et~al.~\cite{coatanea2021} proposed a framework based on dimensionless metrics and singular value decomposition (SVD), demonstrated through LPBF case studies. Zhao~et~al.~\cite{zhao2022} developed an object-oriented information model incorporating a fuzzy–genetic algorithm to represent processing, tooling, and feature data, enabling manufacturability evaluation under resource constraints. Sommers~et~al.~\cite{sommers2024} presented a hybrid rule-based and fuzzy expert system that applies cost-based manufacturability criteria for early-stage design evaluation when detailed CAD data are unavailable. Deep~et~al.~\cite{deep2025} introduced a numerical framework for preliminary manufacturability evaluation in metal powder bed fusion (PBF-LB/M) processes, quantifying manufacturability through measurable geometric and process parameters.

The reviewed studies show that most existing manufacturability assessment frameworks follow geometry-driven pipelines that depend on discretized CAD data, requiring costly preprocessing, causing information loss, increasing training time, and offering limited visibility into how design parameters influence model decisions. The present work advances the field by introducing a parameter-centric, tolerance-aware, and functionally interpretable methodology built on KANs, which learns directly from structured tabular design attributes and enables multi-parameter diagnostics with reversible design feedback. In the following section, the data structure and the synthetic data generation process used in this study are outlined.

\section{Synthetic data generation}
\label{sec4_data_generation}

The performance and effectiveness of DL models are fundamentally governed by the quality of the data that drives them~\cite{Sun2017, bhatt2024}. Numerous studies proved that data quality, diversity, and representativeness remain the most critical determinants of model success~\cite{Recht2019}. In design and manufacturing criteria, well-structured datasets capturing both feasible and non-feasible configurations significantly enhance the predictive accuracy and interpretability of AI models~\cite{khinvasara2024, colombi2024, zha2025}. For manufacturability assessment, balanced sampling across geometric and process-related features enables models to learn from both optimal and failure cases, where understanding why a design fails is as critical as predicting success. Zimmerling~et~al.~\cite{zimmerling2022} showed that data quality influences the performance of DL models for manufacturability assessment more strongly than data quantity, underscoring the importance of high-quality data in geometry-dependent predictions. While some studies focus on cost-based, data-independent manufacturability evaluation~\cite{sommers2024}, this study addresses scenarios where detailed CAD data are available. In this section, we first outline the shortcomings of prevalent data types and data processing methods, and then introduce our novel data structure, dataset characteristics, and the process of data generation.

\subsection{Limitations of CAD-Based Data Processing in Manufacturability Assessment}
\label{sec4_1_CAD_challenges}

As outlined \hyperref[sec3_related_works]{\textcolor{blue}{Section~\ref*{sec3_related_works}}}, extensive research has focused on processing CAD data directly for manufacturability analysis. These methods convert exact CAD representations into grid-compatible tensors by discretizing geometries to extract volumetric information for model training. The most common approach transforms CAD files into volumetric occupancy, or voxel, grids that approximate continuous shapes through uniform cubic sampling~\cite{wu2015shapenets}. Most CAD software, however, stores models in native parametric and B-Rep formats that define geometry analytically through surfaces and topological relations, ensuring precision and computer-aided manufacturing (CAM) compatibility. Since analytic B-Rep structures cannot be directly voxelized, models must first be converted into tessellated polygonal meshes (e.g., STL or OBJ), which approximate continuous geometry using triangular facets. These meshes are then voxelized into 3D occupancy grids that serve as structured volumetric inputs for 3D-CNN architectures in GDL and manufacturability analysis. Each voxel encodes whether a spatial region is occupied (1) or empty (0), although richer encodings such as material attributes or surface normals can also be incorporated. While alternative representations such as point clouds, meshes, and signed distance fields (SDFs) exist, voxel grids remain the most widely used and CNN-compatible format for manufacturing-related GDL tasks. \hyperref[table_voxel_limitations]{\textcolor{blue}{Table~\ref*{table_voxel_limitations}}} summarizes the inherent limitations of voxel-based approaches.

\begin{table*}[!h]
\centering
\caption{Key limitations of voxel-based representations in CAD-driven GDL.} 
\label{table_voxel_limitations}
\begin{tabular}{p{0.84\linewidth} p{0.10\linewidth}}
\hline
Limitations of voxel-based approaches & References \\
\hline
Information loss introduced during CAD file discretization. & \cite{Liu2019} \\
Inability to accurately capture fine geometric details such as micro-holes, thin walls, small fillets, and chamfers. & \cite{Ghadai2021} \\
High sensitivity to resolution, leading to degraded accuracy at lower voxel densities. & \cite{Liu2019,Ghadai2021} \\
Difficulty in incorporating geometric tolerances within rigid voxel grids, necessitating advanced representations such as continuous or probabilistic voxel fields. & \cite{Grosseheide2023,Hilbig2023} \\
Dependence on extensive data augmentation (e.g., 90° rotations) to achieve orientation invariance. & \cite{Fei2024,Mumuni2022} \\
Significant computational cost and prolonged training time due to the large data volume associated with high-resolution voxel tensors. & \cite{Maturana2015,Wu2015} \\
\hline
\end{tabular}
\end{table*}

\begin{figure}[t]
    \centering
    \includegraphics[width=1\linewidth]{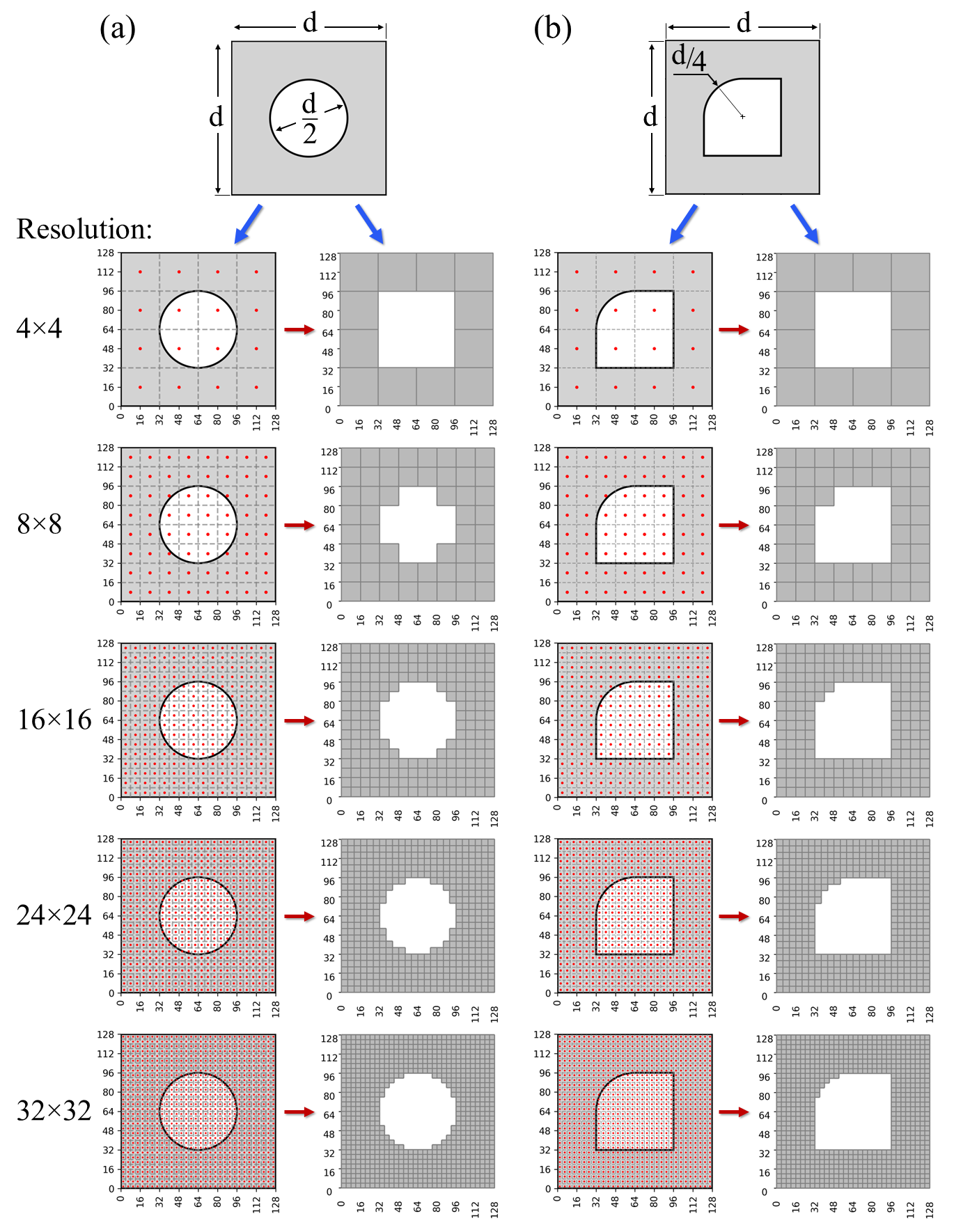} 
    \caption{Effect of descritization resolution on feature representation and information loss for (a) hole with diameter~$\phi = \frac{d}{2}$, (b) pocket with corner radius~$r = \frac{d}{4}$.}
    \label{fig2_resolution_feature}
\end{figure}

\hyperref[fig2_resolution_feature]{\textcolor{blue}{Figure~\ref*{fig2_resolution_feature}}} illustrates how voxelization resolution affects geometric fidelity for holes and fillets and reveals that at least four voxel cells are required to roughly capture the presence of a hole with diameter~$\phi$, whereas sixteen cells or more are needed to represent its size, shape, and boundary curvature accurately. Similarly, representing a pocketwith corner radius~$r$ requires at least four voxels within the fillet region for coarse approximation and eight or more to preserve curvature and boundary smoothness.

\begin{figure}[t]
    \centering
    \includegraphics[width=1\linewidth]{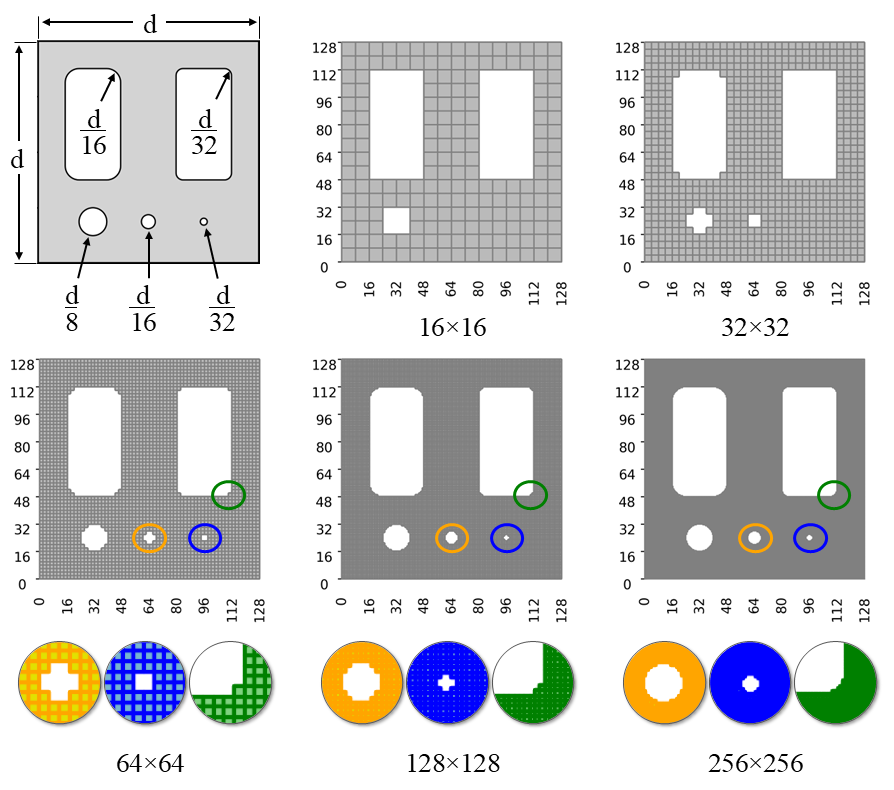}
    \caption{Effect of descritization resolution on feature representation and information loss for a part containing micro features (a hole with diameter~$\phi = \tfrac{1}{32}d$ and a pocket with corner radius~$r = \tfrac{1}{32}d$).}
    \label{fig3_resolution_part}
\end{figure}

\hyperref[fig3_resolution_part]{\textcolor{blue}{Figure~\ref*{fig3_resolution_part}}} further demonstrates the compound effect of feature size and voxelization resolution on geometric fidelity. For a prismatic part with length~$d$, a hole of diameter~$\phi = \frac{1}{16}d$, and a pocket with internal corner radius~$r = \frac{1}{32}d$, a grid resolution of at least $256 \times 256$ is needed to capture both features accurately without distortion. However, industrial designs often include even smaller-scale features, necessitating extremely high resolutions for precise manufacturability analysis. At such fidelity, each design instance generates a cubic voxel tensor comprising more than 16 million elements, substantially increasing preprocessing time and computational cost, which highlights the trade-off between geometric accuracy and data efficiency in voxel-based DL frameworks.

These examples illustrate how limited voxel resolution results in geometric distortion, loss of fine details, and inaccurate spatial representation of small or curved features within voxel-based models. They collectively underscore the need for a more efficient, interpretable, and numeric-based representation of CAD designs, one that preserves geometric and functional fidelity without the inherent drawbacks of voxel-based discretization.

\subsection{Overview of the Proposed Data Structure}
\label{sec4_2_Overview_data}

Building on the critical role of data in deep learning and the limitations of CAD discretization methods, this study introduces a novel data structure that represents designs in a tabular form rather than converting CAD files into volumetric occupancy grids. In this approach, the parameters defining a geometry are compiled into a structured table for direct model training. For example, \hyperref[table1_params_drilling]{\textcolor{blue}{Table~\ref*{table1_params_drilling}}} shows that the overall component dimensions, hole dimensions, and hole locations relative to boundaries can effectively describe a drilling process within a lightweight tabular format. Additional parameters such as material properties, surface roughness, weight, or geometric tolerances can be seamlessly incorporated as new design features. The proposed representation can also be extended to more complex manufacturing scenarios. A single-hole drilling case, for instance, can be expanded to multi-hole configurations by adding the dimensions and locations of additional holes. If holes are placed on different faces, an extra parameter indicating the target face can be included.

When combined with an interpretable DL model such as KAN, this representation identifies the root causes of non-manufacturability and guides designers on which parameters to adjust for feasible, manufacturable designs. The approach enables rapid assessment and is fully reversible, allowing modified parameters to be transformed back into manufacturable CAD files. Although demonstrated on a limited set of scenarios, the framework is readily scalable to diverse geometries, processes, and their combinations, challenges that traditional FR–based methods struggle to address. Not only does this novel approach overcome the inherent limitations of CAD discretization methods discussed in \hyperref[sec4_1_CAD_challenges]{\textcolor{blue}{Section~\ref*{sec4_1_CAD_challenges}}}, but it also offers several distinct advantages:

\begin{enumerate}
    \item It eliminates the need for time-consuming and resource-intensive preprocessing steps, such as converting CAD files to STL format and subsequently voxelizing them.
    \item It enables the seamless incorporation of all types of dimensional and geometric tolerances into the dataset with minimal effort.
    \item It provides significantly higher interpretability compared to geometric-based approaches, offering clear insight into the relationships between design parameters and manufacturability outcomes.
    \item It delivers direct feedback on which design parameters contribute to non-manufacturability.
    \item It maintains full compatibility with the original design language, allowing easy conversion back to manufacturable CAD models.
\end{enumerate}

\subsection{Manufacturing Scenarios and Data Generation}
\label{sec4_3_data_generation}

In this study, three commonly used manufacturing scenarios and their corresponding tabular representations are considered to demonstrate the feasibility of the proposed approach. The scenarios include \textit{hole drilling}, represented by a rectangular block containing a circular hole (\hyperref[table1_params_drilling]{\textcolor{blue}{Table~\ref*{table1_params_drilling}}}); \textit{pocket milling}, represented by a rectangular block containing a rectangular pocket (\hyperref[table2_params_pocket]{\textcolor{blue}{Table~\ref*{table2_params_pocket}}}); and \textit{combined drilling-milling}, represented by a rectangular block containing both a hole and a pocket (\hyperref[table3_params_combined]{\textcolor{blue}{Table~\ref*{table3_params_combined}}}). In each scenario, the component dimensions, manufacturing feature dimensions, and dimensional tolerances are treated as design parameters. These parameters correspond directly to those required to model such geometries in standard CAD software. As discussed in \hyperref[sec4_2_Overview_data]{\textcolor{blue}{Section~\ref*{sec4_2_Overview_data}}}, although additional design parameters can be incorporated, this study limits the attributes to those outlined in \hyperref[table1_params_drilling]{\textcolor{blue}{Table~\ref*{table1_params_drilling}-~\ref*{table3_params_combined}}}.

\begin{table*}[t]
\centering

\begin{minipage}[t]{0.32\textwidth}
\centering
\captionof{table}{Geometric and tolerance parameters for the hole drilling scenario (Scenario~1).}
\label{table1_params_drilling}
\includegraphics[width=\linewidth]{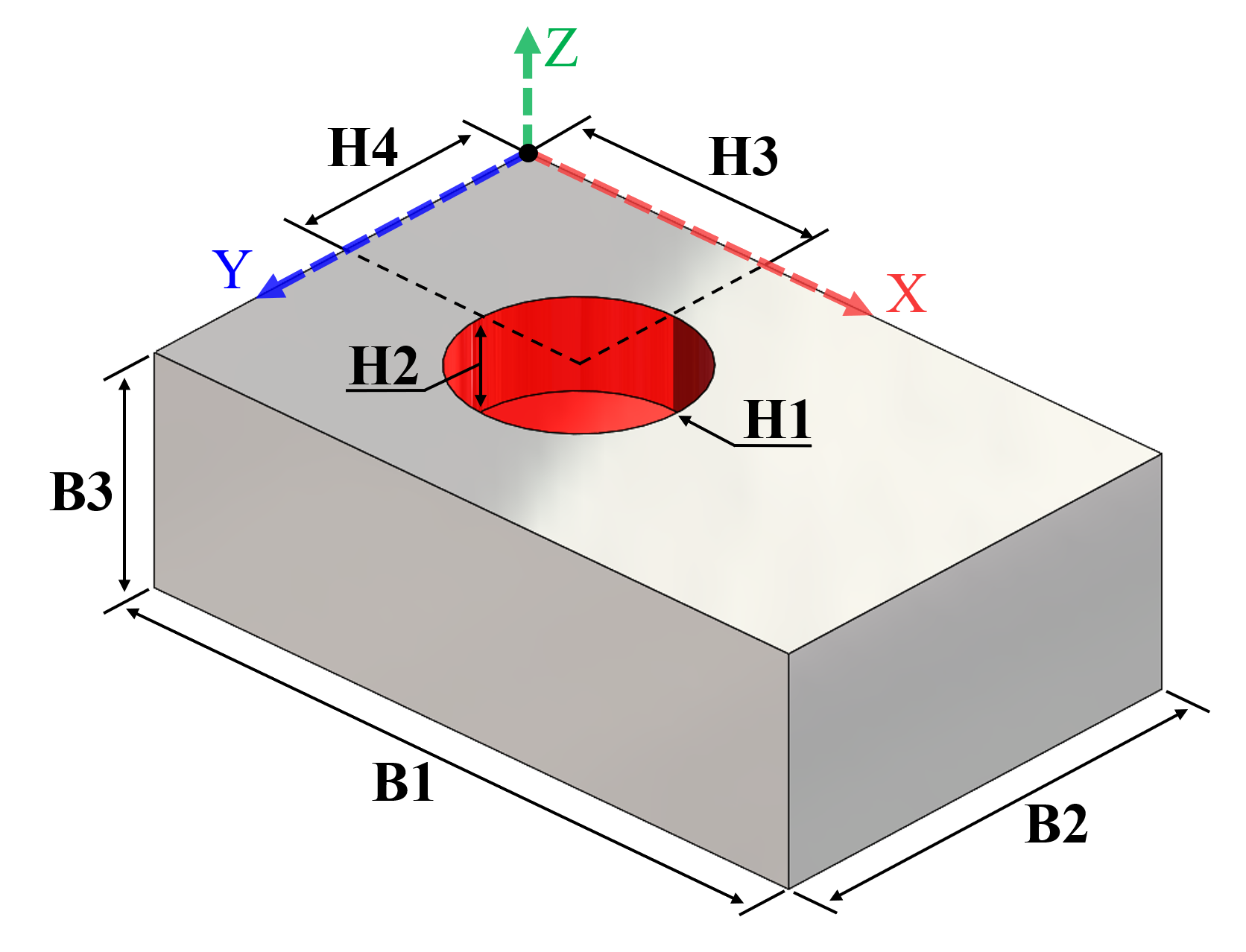}
\\[0.5em]
\small
\begin{tabular}{l l r r}
\toprule
Varaible & Description & \multicolumn{1}{c}{min} & \multicolumn{1}{c}{max} \\
\midrule
B1    & Block length     & 15.0  & 200.0 \\
B2    & Block width      & 14.0  & 199.3 \\
B3    & Block height     &  4.0  & 200.0 \\
H1    & Hole diameter    &  1.0  &  30.0 \\
H2    & Hole depth       &  4.0  & 198.9 \\
H3    & Hole x pos.      &  1.5  & 196.2 \\
H4    & Hole y pos.      &  1.5  & 177.6 \\
H1\_UT & H1 Upper Tol.   & -0.50  &   0.50 \\
H1\_LT & H1 Lower Tol.   & -0.50  &   0.50 \\
H2\_UT & H2 Upper Tol.   & -0.50  &   0.50 \\
H2\_LT & H2 Lower Tol.   & -0.50  &   0.50 \\
H3\_UT & H3 Upper Tol.   & -0.50  &   0.50 \\
H3\_LT & H3 Lower Tol.   & -0.50  &   0.50 \\
H4\_UT & H4 Upper Tol.   & -0.50  &   0.50 \\
H4\_LT & H4 Lower Tol.   & -0.50  &   0.50 \\
{}     &        {}       &   {}   &    {}   \\
{}     &        {}       &   {}   &    {}   \\
\bottomrule
\end{tabular}
\end{minipage}
\hfill
\begin{minipage}[t]{0.32\textwidth}
\centering
\captionof{table}{Geometric and tolerance parameters for the pocket milling scenario (Scenario~2).}
\label{table2_params_pocket}
\includegraphics[width=\linewidth]{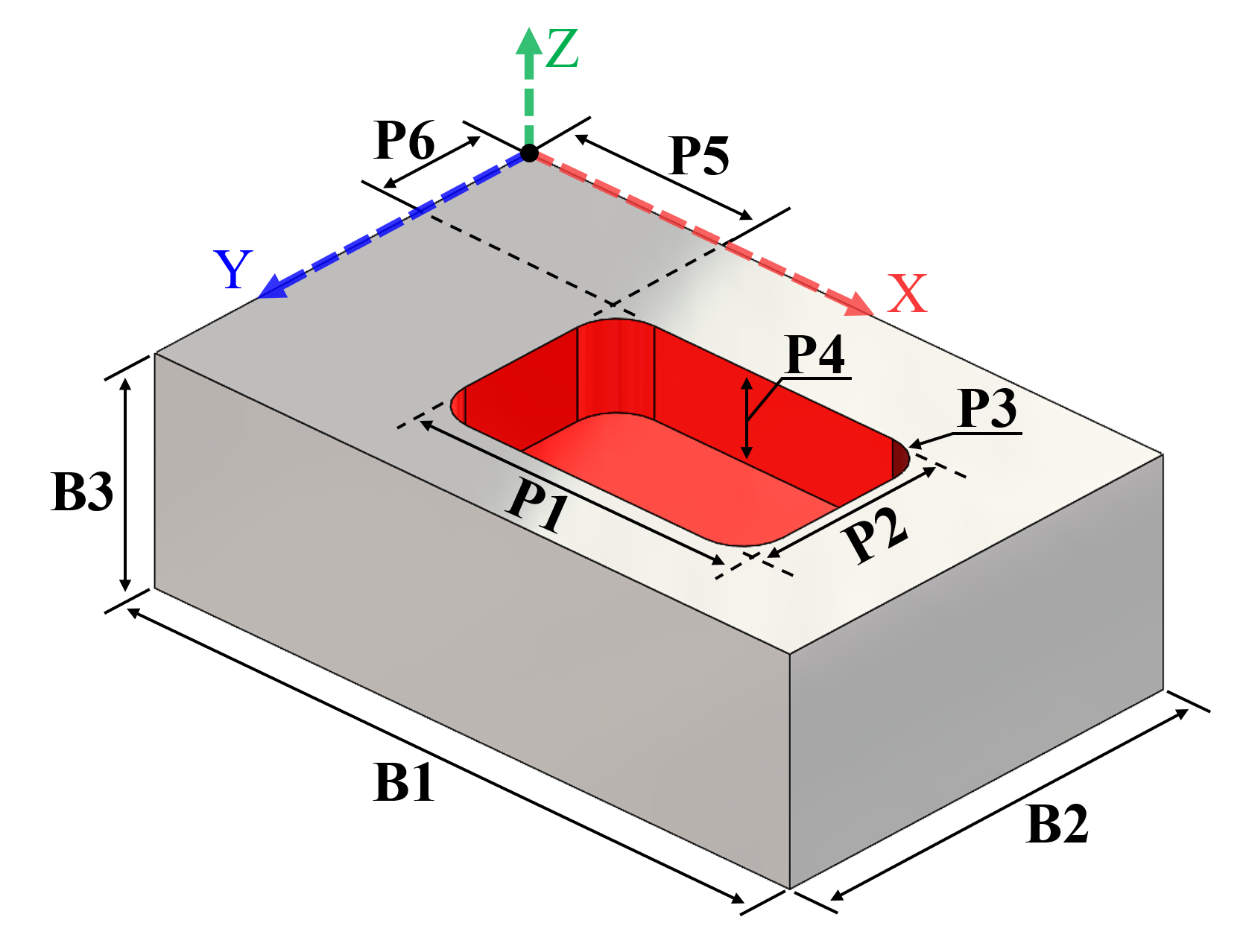}
\\[0.5em]
\small
\begin{tabular}{l l r r}
\toprule
Varaible & Description & \multicolumn{1}{c}{min} & \multicolumn{1}{c}{max} \\
\midrule
B1--B3 & As in \hyperref[table1_params_drilling]{\textcolor{blue}{Table~\ref*{table1_params_drilling}}} & ---    & ---    \\
P1     & Pocket length      &  3.4   & 193.7 \\
P2     & Pocket width       &  3.0   & 180.0 \\
P3     & Pocket radius      &  0.0   &  25.0 \\
P4     & Pocket depth       &  2.0   & 195.2 \\
P5     & Pocket x pos.      &  0.6   & 182.0 \\
P6     & Pocket y pos.      &  0.6   & 155.8 \\
P1\_UT & P1 Upper Tol.      & -0.50  &   0.50 \\
P1\_LT & P1 Lower Tol.      & -0.50  &   0.50 \\
P2\_UT & P2 Upper Tol.      & -0.50  &   0.50 \\
P2\_LT & P2 Lower Tol.      & -0.50  &   0.50 \\
P4\_UT & P4 Upper Tol.      & -0.50  &   0.50 \\
P4\_LT & P4 Lower Tol.      & -0.50  &   0.50 \\
P5\_UT & P5 Upper Tol.      & -0.50  &   0.50 \\
P5\_LT & P5 Lower Tol.      & -0.50  &   0.50 \\
P6\_UT & P6 Upper Tol.      & -0.50  &   0.50 \\
P6\_LT & P6 Lower Tol.      & -0.50  &   0.50 \\
\bottomrule
\end{tabular}
\end{minipage}
\hfill
\begin{minipage}[t]{0.32\textwidth}
\centering
\captionof{table}{Geometric and tolerance parameters for the combined drilling--milling scenario (Scenario~3).}
\label{table3_params_combined}
\includegraphics[width=\linewidth]{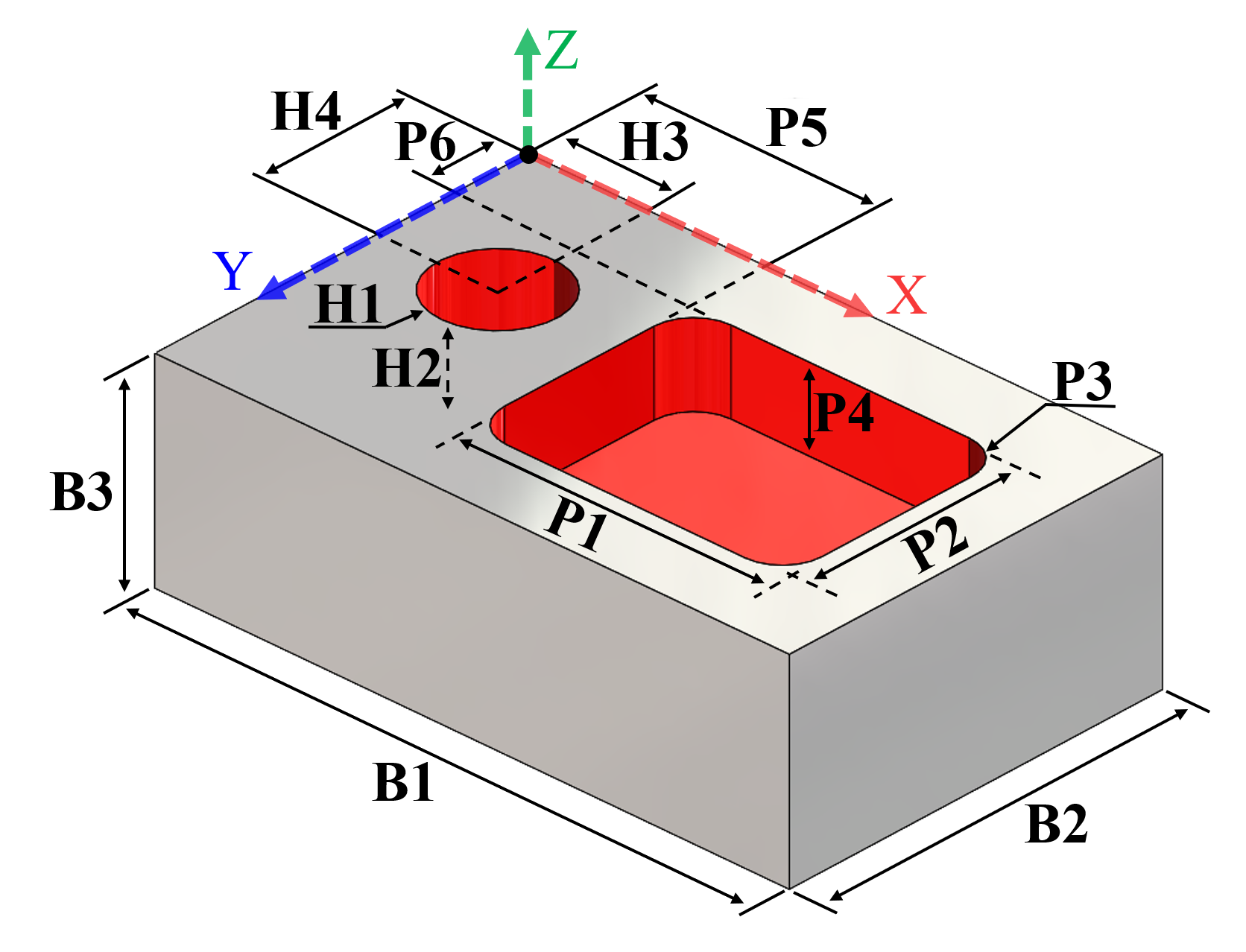}
\\[0.5em]
\small
\begin{tabular}{l l r r}
\toprule
Varaible & Description & \multicolumn{1}{c}{min} & \multicolumn{1}{c}{max} \\
\midrule
B1     & Block length    & 25.3 & 300.0 \\
B2     & Block width     & 16.2 & 296.7 \\
B3     & Block height    &  4.0 & 299.9 \\
H1     & Hole diameter   &  1.0 &  30.0 \\
H2     & Hole depth      &  2.0 & 297.1 \\
H3     & Hole x pos.     &  1.7 & 294.0 \\
H4     & Hole y pos.     &  1.6 & 289.8 \\
P1     & Pocket length   &  5.8 & 150.0 \\
P2     & Pocket width    &  3.2 & 148.5 \\
P3     & Pocket radius   &  0.0 &  25.0 \\
P4     & Pocket depth    &  2.0 & 294.7 \\
P5     & Pocket x pos.   &  0.6 & 284.2 \\
P6     & Pocket y pos.   &  0.6 & 270.4 \\
H\_Tol. & As in \hyperref[table1_params_drilling]{\textcolor{blue}{Table~\ref*{table1_params_drilling}}} & ---   & ---   \\
P\_Tol. & As in \hyperref[table2_params_pocket]{\textcolor{blue}{Table~\ref*{table2_params_pocket}}}   & ---   & ---   \\
{}     &        {}       &   {}   &    {}   \\
{}     &        {}       &   {}   &    {}   \\
\bottomrule
\end{tabular}
\end{minipage}

\end{table*}

To properly investigate the effect of data size on model performance, a synthetic dataset containing 100,000 labeled designs was generated for each of the scenarios mentioned above. The datasets were algorithmically created to represent both feasible and non-feasible configurations across the aforementioned manufacturing scenarios. Each dataset was generated using a Python-based parametric case-study generator developed by the authors. This script encodes manufacturing constraints and DFM rules to label each instance as manufacturable or non-manufacturable automatically.

In this study, manufacturability is strictly defined by design and machining feasibility derived from established DFM standards, reference literature, and textbooks, and is confined to two primary subtractive machining operations: milling and drilling. The manufacturability criteria encoded during data synthesis are summarized in \hyperref[table5_manufac_criteria]{\textcolor{blue}{Table~\ref*{table5_manufac_criteria}}}. Non-manufacturability may arise from geometric infeasibility (e.g., sharp internal pocket corners), structural constraints (e.g., insufficient remaining wall or floor thickness affecting part function or rigidity), or tooling limitations. For example, a pocket geometry may satisfy DFM rules yet remain impossible to machine when no standard end mill exists with the required diameter and flute length to cut a cavity of that depth and internal corner radius.

\begin{table}[h]
    \centering
    \caption{Manufacturing processes and associated manufacturability criteria.}
    \begin{tabular}{p{0.15\linewidth} p{0.48\linewidth} p{0.20\linewidth}} 
    \hline
        Process & Manufacturability criteria & References \\
    \hline
        {} & Twist drill size range & \\
        {} & Depth-to-diameter ratio & \\
        Hole & Hole positioning and spacing & \cite{msc2019, machinery2020_1, groover_ch22, bralla_ch4_5, sandvik_drilling, asm1989_drill, kalpakjian2021_ch23, boothroyd2010_drill, machinery2020_2, iscar2024,bralla_ch3_7, bralla_ch6_10, iso2768_1_1989}  \\
        drilling & Tool reachability & \\
        {} & Tolerances & \\

    \hline
        {} & End mill size range &  \\
        {} & Corner radii &  \\
        Pocket & Depth-to-tool diameter ratio & \cite{machinery2020_3, kalpakjian2021_ch23, kalpakjian2021_ch24, boothroyd2010_mill, msc2019}\\
        milling & Wall and bottom thickness & \cite{groover_ch23, bralla_ch1_4, asm1989_mill,iso2768_1_1989}\\
        {} & Tool reachability & \\
        {} & Tolerances & \\
    \hline
    \end{tabular}
    \label{table5_manufac_criteria}
\end{table}

To reflect these often-overlooked constraints, this work jointly embeds commercial tool availability and DFM feasibility rules into the data generation process, producing a more realistic and industry-relevant dataset. A total of 20 unique non-manufacturability conditions are modeled for \textit{hole drilling}, 20 for \textit{pocket milling}, and 252 for \textit{combined drilling-milling}, capturing both individual process constraints and their interactions. Within each dataset, 50\% of case studies are generated randomly across the defined parameter ranges, while the remaining 50\% are deliberately created near the DFM boundary conditions, where slight variations in dimensions or tolerances can alter manufacturability status. This sampling strategy challenges the DL model to evaluate manufacturability under extreme design and process conditions. For other machining processes or manufacturing features, corresponding constraints can be integrated into the same script to generate new parametric datasets. The dimensional ranges considered in this study were selected to ensure that all manufacturable designs can be fabricated in a prototyping machine shop. However, the framework can be easily scaled to industrial-level applications. The generator script is designed to produce a balanced dataset in which manufacturable and non-manufacturable cases are equally represented.

Building upon the generated synthetic datasets, the following section presents the proposed manufacturability assessment methodology. It provides a detailed description of the KAN-based network architecture, the training and evaluation procedures, and the interpretability tools employed to analyze model behavior and parameter influence.

\section{Methodology}
\label{sec5_methodology}

\subsection{Overview of the Proposed Approach}
\label{sec5_1_method_overview}

\begin{figure}[t]
    \centering
    \includegraphics[width=1\linewidth]{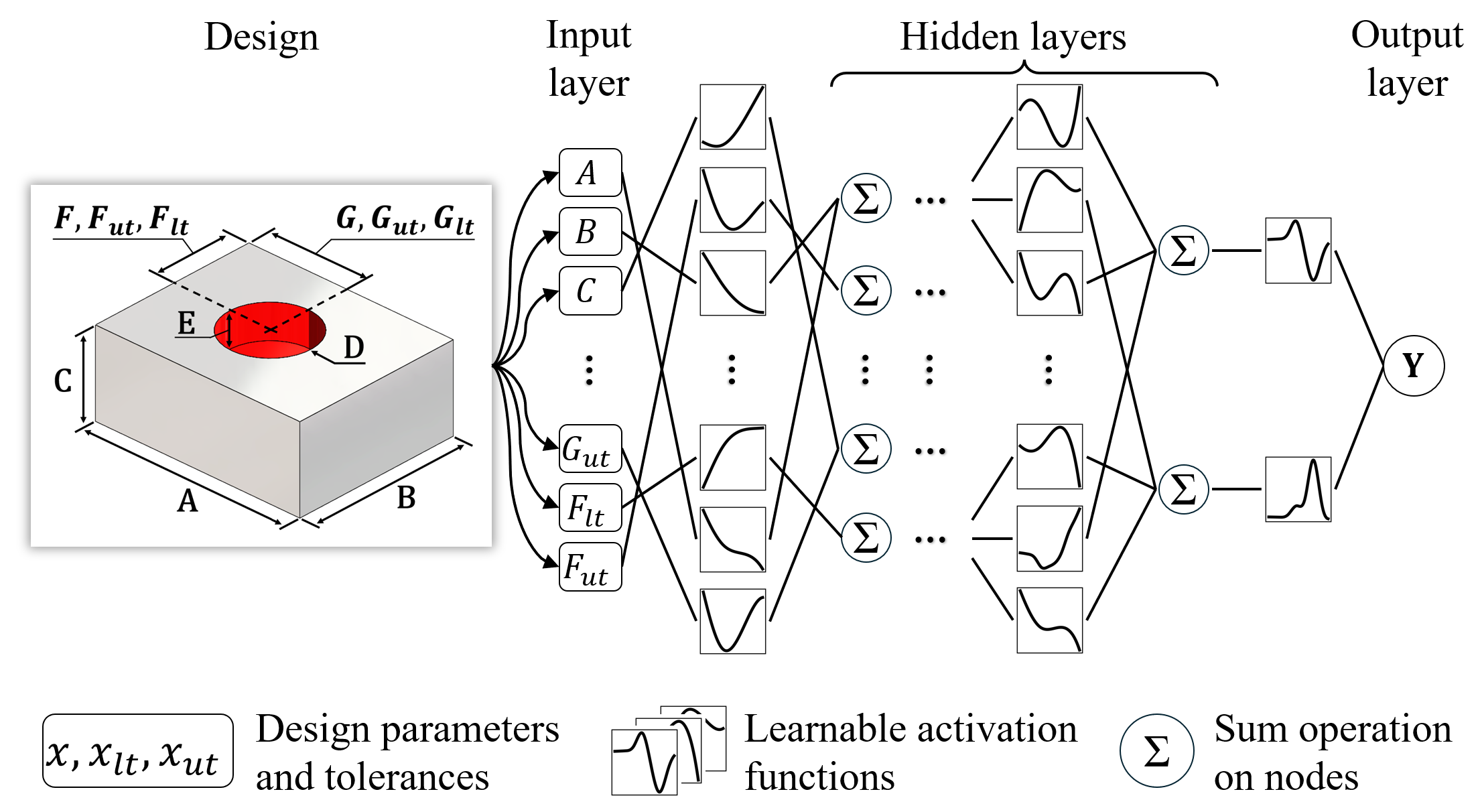}
    \caption{General structure of the proposed manufacturability analyzer model.}
    \label{fig4_KAN_arch}
\end{figure}

This study presents an interpretable methodology for manufacturability evaluation that moves beyond conventional geometry-based DL paradigms by formulating manufacturability as a parameter-based problem rather than relying on direct geometric learning. As illustrated in \hyperref[fig4_KAN_arch]{\textcolor{blue}{Figure~\ref*{fig4_KAN_arch}}}, the proposed model is built on KANs~\cite{liu2024kan}, which capture the nonlinear mapping between design parameters and manufacturability labels, enabling parameter-level interpretability. Parametric design features and their associated tolerances serve as inputs to the model, which performs binary classification to determine whether the given set of parameters corresponds to a manufacturable or non-manufacturable design. Beyond prediction, the methodology provides diagnostic insights into the causes of non-manufacturability and quantifies how individual parameters influence classification outcomes. Latent space projections and SHapley Additive exPlanations (SHAP) analysis are further employed to interpret the learned feature space and identify parameter combinations most responsible for non-manufacturable cases. Overall, the framework transforms manufacturability prediction from a black-box process into an interpretable, data-driven reasoning system that directly links design intent with manufacturability feedback.

\subsection{Kolmogorov–Arnold Networks}
\label{sec5_2_kan}

The proposed methodology relies on Kolmogorov--Arnold Networks~\cite{liu2024kan}, a recently introduced class of neural networks that are directly inspired by the Kolmogorov--Arnold representation theorem~\cite{kolmogorov1957representation, kolmogorov1961representation}. Based on this theorem, any continuous multivariate function can be represented precisely as a finite superposition of univariate functions and addition operators. In its classical form, the theorem can be stated as follows:

Let $f : [0,1]^n \rightarrow \mathbb{R}$ be a continuous function. Then there exist continuous univariate functions
\[
\phi_q : \mathbb{R} \rightarrow \mathbb{R}, 
\quad 
\psi_{p,q} : [0,1] \rightarrow \mathbb{R},
\]
for $p = 1,\dots,n$ and $q = 1,\dots,2n+1$, such that
\begin{equation}
    f(x_1,\dots,x_n) 
    = \sum_{q=1}^{2n+1} 
    \phi_q\!\left(
        \sum_{p=1}^{n} \psi_{p,q}(x_p)
    \right),
    \label{eq_kolmogorov_representation}
\end{equation}
for all $(x_1,\dots,x_n) \in [0,1]^n$.

In another words, any continuous $n$-variate function can be expressed as a finite sum of univariate ``outer'' functions $\phi_q$ applied to linear combinations of univariate ``inner'' functions $\psi_{p,q}$ acting on each input dimension. This result provides a constructive superposition framework in which multivariate complexity is captured through compositions and sums of simpler one-dimensional functions.

KANs operationalize the representation in \hyperref[eq_kolmogorov_representation]{\textcolor{blue}{Eq.~(\ref*{eq_kolmogorov_representation})}} by replacing fixed affine weights in standard neural networks with learnable univariate functions parameterized by splines~\cite{liu2024kan}. Considering a network layer with input vector $\mathbf{x}^{(\ell)} \in \mathbb{R}^{d_\ell}$ and output vector $\mathbf{x}^{(\ell+1)} \in \mathbb{R}^{d_{\ell+1}}$, the mapping in a KAN layer is defined as:
\begin{equation}
    x^{(\ell+1)}_j 
    = \sum_{i=1}^{d_\ell} \phi^{(\ell)}_{ij}\!\big(x^{(\ell)}_i\big),
    \quad j = 1,\dots,d_{\ell+1},
    \label{eq:kan_layer}
\end{equation}
where each $\phi^{(\ell)}_{ij} : \mathbb{R} \rightarrow \mathbb{R}$ is a learnable univariate function associated with the connection (edge) from neuron $i$ in layer $\ell$ to neuron $j$ in layer $\ell+1$. The edge itself is a function and no fixed linear weight $w_{ij}$ appears explicitly.

To make this architecture trainable and numerically stable, each edge function $\phi^{(\ell)}_{ij}$ is parameterized using a spline basis. For example, using B-splines of order $k$ over a grid of knots $\{t_m\}_{m=1}^{M}$, yields:
\begin{equation}
    \phi^{(\ell)}_{ij}(x)
    = \sum_{m=1}^{M} c^{(\ell)}_{ijm} \, B_{m}^{(k)}(x;\,\mathbf{t}),
    \label{eq_phi_spline}
\end{equation}
where $B_{m}^{(k)}(\cdot;\mathbf{t})$ denotes the $m$-th B-spline basis function of order $k$ defined over the knot vector $\mathbf{t}$, and $c^{(\ell)}_{ijm}$ are learnable coefficients. The overall network implements a composition of such spline-based edge functions and summation operations across layers, forming a flexible approximation of the target mapping:
\[
    f : \mathbf{x} \mapsto \hat{p},
\]
where, in this work, $\mathbf{x}$ denotes the vector of design parameters and $\hat{p} \in [0,1]$ represents the predicted probability that a design is manufacturable.

For evaluation, probabilistic outputs are converted to class labels using a decision threshold $\tau$ (set to $\tau = 0.5$ in this study):
\[
\hat{y} =
\begin{cases}
1, & \hat{p} \ge \tau,\\[3pt]
0, & \hat{p} < \tau.
\end{cases}
\]
Designs with $\hat{p} \ge 0.5$ are therefore classified as manufacturable, and those with $\hat{p} < 0.5$ as non-manufacturable. In applications requiring more conservative decisions, this threshold may be increased (e.g., $\tau > 0.5$) to ensure that only high-confidence predictions are accepted as manufacturable.

Given a set of design parameters $\mathbf{x} \in \mathbb{R}^d$ and corresponding binary manufacturability labels $y \in \{0,1\}$, the KAN is trained using the standard binary cross-entropy loss:
\begin{equation}
    \mathcal{L}(\Theta)
    = - \frac{1}{N} \sum_{n=1}^{N} 
    \left[
        y_n \log \hat{p}_n 
        + (1 - y_n)\log (1 - \hat{p}_n)
    \right]
    + \lambda \, \mathcal{R}(\Theta),
    \label{eq_kan_loss}
\end{equation}
where $\hat{p}_n$ is the predicted manufacturability probability for the $n$-th sample, $\Theta$ denotes the collection of all spline coefficients and grid parameters, $\mathcal{R}(\Theta)$ is a regularization term, and $\lambda$ is its weighting coefficient.

A key advantage of KANs in the context of manufacturability assessment is their intrinsic interpretability. Because the network is constructed from explicit univariate functions along each edge, it is possible to inspect the learned functions $\phi^{(\ell)}_{ij}(x)$ directly as one-dimensional curves, revealing how each intermediate neuron responds to variations in specific input or latent coordinates, compute and visualize partial derivatives $\partial f / \partial x_i$ via the chain rule, leveraging the analytic derivatives of the spline basis functions in \hyperref[eq_phi_spline]{\textcolor{blue}{Eq.~(\ref*{eq_phi_spline})}}, and relate changes in individual design parameters to changes in the manufacturability output, thereby connecting human-interpretable design parameters to model predictions.

\subsection{Data Preprocessing}
\label{sec5_3_data_process}

As outlined in \hyperref[sec4_data_generation]{\textcolor{blue}{Section~\ref*{sec4_data_generation}}}, three manufacturing scenarios each containing 100,000 distinct case studies are considered in this study to showcase the proposed manufacturability assessment approach. Prior to model training and to ensure that all design parameters contribute proportionally to the learning process and to prevent features with large numerical ranges from dominating model training, all continuous parameters were normalized to the range $[0,1]$ using the Min--Max scaling transformation:

\begin{equation}
x' = \frac{x - x_{\min}}{x_{\max} - x_{\min}},
\label{eq:minmax}
\end{equation}

\noindent where $x$ is unscaled value of a design parameter, $x_{\min}$ and $x_{\max}$ are the minimum and maximum values of that parameter within the training set, and $x'$ is the normalized value scaled to the range $[0,1]$. Categorical variables were transformed into binary indicator vectors using one--hot encoding. For a categorical feature with $K$ unique categories $\{c_1, c_2, \dots, c_K\}$, each sample is represented as a $K$-dimensional binary vector $\mathbf{z} = [z_1, z_2, \dots, z_K]$, where:

\begin{equation}
z_k =
\begin{cases}
1, & \text{if the sample belongs to category } c_k,\\
0, & \text{otherwise.}
\end{cases}
\label{eq:onehot}
\end{equation}

This encoding converts discrete design characteristics into a numerical format compatible with the KAN model while preventing the algorithm from inferring ordinal relationships among categorical values. Together, Min--Max scaling and one--hot encoding produce a balanced and well-structured feature space, allowing the KAN to accurately learn the nonlinear mapping between design parameters and manufacturability outcomes.

The processed data were randomly shuffled and split into training (80\%) and testing (20\%) sets to ensure statistical consistency across experiments. This representation provides a compact yet comprehensive numerical description of each design, preserving the essential geometric and tolerance information necessary for manufacturability assessment while ensuring compatibility with KANs.

\subsection{Training and Evaluation Procedure}
\label{sec5_4_train_eval}

In this study, KANs were implemented using the \textit{PyKAN} package~\cite{liu2024kan, pykan2024}, which provides a flexible and practical realization of the architecture mentioned above. The primary hyperparameters of the KAN model include the architecture of the hidden layers (i.e., the number of spline-activation neurons per layer), the number of grid points that define the resolution of each learnable spline function, the spline order that controls the smoothness and expressive capacity of the activation functions, the choice of optimization algorithm used to update the spline coefficients and weights, and the number of training iterations performed during optimization.

To investigate data efficiency and determine the minimum data requirement for stable learning, a learning curve analysis was conducted for each machining scenario. The model was trained iteratively using incremental subsets ranging from 1,000 to 100,000 samples, while maintaining identical validation and testing partitions. Performance metrics were recorded at each increment to observe how predictive accuracy and AUC evolved with increasing data size. The results indicated that performance stabilized beyond 30,000 samples for the \textit{hole drilling} and \textit{pocket milling} scenarios, and beyond 40,000 samples for the \textit{milling–drilling} scenario. These findings informed the final dataset sizes used for complete training and testing, ensuring computational efficiency without sacrificing predictive fidelity.

A systematic brute-force hyperparameter search was conducted to determine the most reliable and high-performing KAN configuration for the manufacturability prediction task. The search explored a broad range of architectural and spline-related settings to capture models of varying depth, width, and representational capacity. Architectures spanning from shallow networks with a single hidden layer to deeper configurations with up to four hidden layers were examined, with layer widths ranging from compact structures with 8 neurons to more expressive designs with 64 neurons, while keeping the last hidden layer at 2 neurons for latent space visualizations. In parallel, multiple spline resolutions with grid sizes ranging from 3 to 7 and spline orders ($k$) ranging from 2 to 4 were assessed to vary the smoothness and flexibility of the learned activation functions. Each candidate configuration was trained using both a first-order optimizer \textit{(Adam)} and a quasi-Newton optimizer \textit{(LBFGS)}, and evaluated under a consistent early-stopping protocol to ensure fair comparison across settings. This exhaustive exploration enabled a robust assessment of the KAN design space and ensured that the final selected model achieved a favorable balance among predictive accuracy, training stability, and generalization performance.

The tuning procedure explored all parameter combinations stated above, yielding 900 unique configurations per manufacturing scenario. Each configuration was trained and validated using 10-fold stratified cross-validation to ensure generalizability and reduce variance due to data partitioning. The mean validation performance across folds was used to rank configurations. The optimal configuration of \textit{hidden layers} = [16,2], \textit{grid} = 3, \textit{k} = 3, and \textit{optimizer} = LBFGS was determined. The LBFGS optimizer was selected for its quasi-Newton convergence properties, which are well-suited for continuous functional approximation tasks.

The output layer of the network produces a single probability value corresponding to the predicted likelihood of manufacturability, which is thresholded at 0.5 to yield binary classification results. During training, convergence was monitored through validation loss, and the process was regulated using early stopping and learning-rate scheduling strategies. Specifically, the \textit{EarlyStopping} criterion was applied with a patience of 20 training steps, while the \textit{ReduceLROnPlateau} scheduler reduced the learning rate by a factor of 0.5 after three steps without improvement, with a minimum learning rate of $1\times10^{-6}$. These mechanisms ensured stable convergence and prevented overfitting.

Model performance was quantitatively evaluated using multiple complementary metrics. The \textit{Area Under the Receiver Operating Characteristic Curve (AUC)} was chosen as the primary metric, and the \textit{F1-score} as the secondary metric. AUC was prioritized because it provides a threshold-independent measure of class separability, reflecting the model’s ability to distinguish between manufacturable and non-manufacturable designs under varying decision boundaries. This property is especially valuable for engineering classification problems, where the trade-off between false positives (FP) and false negatives (FN) may shift depending on design tolerance or process constraints. The F1-score was used as a complementary measure to capture the balance between precision and recall, ensuring that both error types were effectively managed. Together, these metrics provide a rigorous and interpretable assessment of the model’s predictive reliability.

\rowcolors{2}{gray!10}{white}
\begin{table*}[t]
\centering
\small
\caption{Summary of benchmark models and key hyperparameter configurations used for evaluation.}
\label{table6_models_hyperparams}
\renewcommand{\arraystretch}{1.1}
\begin{tabular}{p{0.15\textwidth} p{0.65\textwidth}}
\toprule
Model & Key hyperparameters \\
\midrule

MLP \cite{rumelhart1986} &
Architecture $[256,128,64]$, learning rate $10^{-2}$ (adaptive), ReLU activation, Adam optimizer \\

FNN \cite{bishop1995} &
Architecture $[128,32]$, dropout $0.2$, learning rate $10^{-3}$, Adam optimizer \\

XGBoost \cite{chen2016xgboost} &
500 trees, max depth $9$, learning rate $0.1$, log-loss objective \\

LightGBM \cite{ke2017lightgbm} &
300 trees, max depth $9$, learning rate $0.1$, 63 leaves \\

GBDT \cite{friedman2001gbm} &
200 trees, max depth $5$, learning rate $0.1$, subsample $0.8$ \\

CatBoost \cite{prokhorenkova2018catboost} &
200 iterations, depth $9$, learning rate $0.1$ \\

ET \cite{geurts2006extratrees} &
300 trees, fully randomized splits, unlimited depth \\

RF \cite{breiman2001rf} &
300 trees, bootstrap disabled, unlimited depth \\

DT \cite{breiman1984cart} &
Max depth $15$, log-loss criterion, cost-complexity pruning \\

KNN \cite{cover1967nn} &
100 neighbors, Minkowski distance ($p=3$), distance weighting \\

RC \cite{hoerl1970ridge} &
$\ell_2$ regularization, $\alpha = 10^{-2}$, LSQR solver \\

LR \cite{cox1958} &
$\ell_2$ regularization, $C=1$, liblinear solver \\

SVM \cite{cortes1995} &
RBF kernel, $C=10$, scale-based $\gamma$ \\

GNB \cite{hand2001} &
Variance smoothing $= 0.1$ \\

\bottomrule
\end{tabular}
\end{table*}

For comparative assessment, a comprehensive set of benchmark models, encompassing conventional ML algorithms and NN architectures, was trained and evaluated under identical experimental conditions, using the same training–testing splits. These benchmarks included MLP, FNN, support vector machines (SVM), k-nearest neighbors (kNN), logistic regression (LR), ridge classifiers (RC), Gaussian naive Bayes (GNB), decision trees (DT), random forests (RF), extremely randomized trees (ET), gradient boosting decision trees (GBDT), XGBoost, LightGBM, and CatBoost. All models employed identical preprocessing steps, and their respective hyperparameters were systematically tuned across predefined search spaces. For each model, the optimal hyperparameter configuration was selected based on validation AUC and F1-score metrics, ensuring that each algorithm was evaluated under its best achievable settings. The evaluated models and their corresponding optimal hyperparameters are summarized in \hyperref[table6_models_hyperparams]{\textcolor{blue}{Table~\ref*{table6_models_hyperparams}}}. This benchmarking procedure established a consistent and equitable basis for comparison, enabling a fair assessment of the proposed Kolmogorov–Arnold Network (KAN) relative to established learning approaches and highlighting its advantages in modeling nonlinear, high-dimensional relationships for manufacturability prediction.

\subsection{Interpretability and Visualization}
\label{sec5_5_interpret_tools}

In addition to focusing on predictive accuracy, the proposed methodology emphasizes interpretability as a core requirement for manufacturability assessment. Understanding the contribution of individual design parameters to model decisions is essential for providing actionable feedback to designers and for validating learned relationships against established DFM principles. To this end, multiple complementary interpretability techniques were employed to analyze both the internal structure of the KAN model and the relationships it learned among features.

First, the intrinsic interpretability of the KAN architecture was exploited through direct visualization of the learned spline functions and their gradients. Each spline represents a univariate nonlinear transformation learned by the network, allowing the influence of individual parameters on manufacturability predictions to be explicitly examined. These plots provide a functional perspective on parameter sensitivity and local monotonicity, revealing which geometric or tolerance features contribute most strongly to manufacturability outcomes.

To further quantify feature importance, the SHAP framework was applied to the trained KAN model. SHAP values were computed for each feature to estimate the average marginal contribution of each feature to the predicted output across the dataset. This analysis identified the dominant design variables and their interactions responsible for non-manufacturable classifications, thereby linking the model’s mathematical reasoning with engineering intuition.

Additionally, three dimensionality-reduction techniques were employed to project the high-dimensional design parameters into a two-dimensional space to inspect potential clustering patterns and the separation between manufacturable and non-manufacturable samples. The first is a KAN-driven latent projection, obtained directly from the two neurons in the final hidden layer of the trained network. These neurons constitute the model’s internal two-dimensional representation of the design space, capturing the nonlinear interactions and manufacturability-relevant relationships learned during training. The second category includes two widely used manifold-learning techniques, including \textit{Uniform Manifold Approximation and Projection (UMAP)} and \textit{t-Distributed Stochastic Neighbor Embedding (t-SNE)}, used to evaluate how conventional nonlinear embeddings represent the structure of the manufacturability landscape.

The third technique is a \textit{feed-forward autoencoder} with a two-neuron bottleneck layer that learns a compact neural latent representation by reconstructing normalized feature inputs. The encoder compresses the input through layers of 64 and 32 neurons before reaching the two-dimensional bottleneck, and the decoder mirrors this structure through layers of 32 and 64 neurons to recover the original feature vector. The model is trained end-to-end using \textit{mean-squared error (MSE)} loss and the \textit{Adam} optimizer, with \textit{EarlyStopping} and \textit{ReduceLROnPlateau} used to prevent overfitting and stabilize convergence. After training, the encoder is used to generate the two-dimensional latent projections.

Collectively, these projections provide a multi-perspective view of how different models organize the design space and illustrate the relative ability of each technique to separate manufacturable and non-manufacturable cases. When combined with spline-function visualization and SHAP-based feature attribution, the full interpretability pipeline offers a comprehensive understanding of how the KAN encodes geometric parameters, tolerances, and their interactions during decision-making. These tools not only expose the internal functional structure of the model but also provide practical, design-centered diagnostics that reveal which parameters drive manufacturability outcomes.

Together, these interpretive capabilities complement the predictive strength of the proposed methodology and establish a unified framework for both performance evaluation and engineering insight. Building on this foundation, the following section presents the quantitative and qualitative results, detailing the KAN’s training behavior, classification performance across scenarios, comparative benchmarking against alternative ML and DL models, and interpretability findings that validate the approach's practical effectiveness.

\section{Experiments and results}
\label{sec6_results}

This section presents the experimental evaluation of the proposed KAN-based manufacturability assessment method across the three aforementioned manufacturing scenarios. The experiments are designed to assess the predictive performance of the KAN model, compare its results with benchmark algorithms outlined in \hyperref[table6_models_hyperparams]{\textcolor{blue}{Table~\ref*{table6_models_hyperparams}}}, and analyze its interpretability using visual and attribution-based techniques.

\subsection{Experimental Setup}
\label{sec6_1_setup}

All experiments were conducted on a Dell Precision 7960 Tower workstation equipped with an Intel\textsuperscript{\textregistered} Xeon\textsuperscript{\textregistered} w9-3475X CPU (72 cores, 2.21 GHz), 256 GB DDR5 RAM (4800 MT/s), and an NVIDIA RTX 4500 Ada Generation GPU (24 GB VRAM). The computational framework was implemented in Python 3.9.23 using the PyTorch 1.12.1 DL library with CUDA 11.3 and cuDNN 8.3.2 for GPU acceleration. The KAN models were implemented using the open-source PyKAN 0.1.8 package ~\cite{pykan2024}. All models were trained and evaluated under identical software configurations to ensure experimental consistency and reproducibility.

\subsection{Scenario-Specific Performance Evaluation}
\label{sec6_2_scenario_eval}

\begin{figure*}[!h]
    \centering
    \includegraphics[width=1\linewidth]{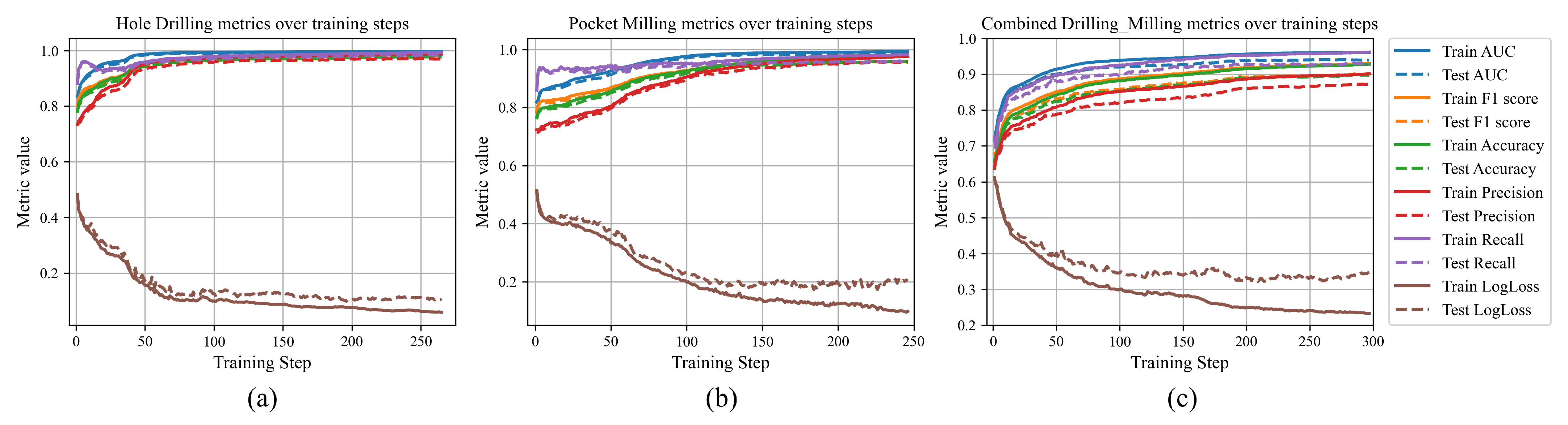}
    \caption{The KAN model evolution of performance metrics for (a) Hole Drilling, (b) Pocket Milling, and (c) Combined Drilling–Milling scenarios.}
    \label{fig4_train_metrics}
\end{figure*}

\hyperref[fig4_train_metrics]{\textcolor{blue}{Figure~\ref*{fig4_train_metrics}}} illustrate the evolution of key performance metrics for all three manufacturing scenarios throughout the KAN training process. For the \textit{hole drilling} scenario trained on 30,000 case studies (\hyperref[fig4_train_metrics]{\textcolor{blue}{Figure~\ref*{fig4_train_metrics}(a)}}), both training and testing AUC rise rapidly, exceeding 0.992 and continuing toward near-perfect class separability. F1-score, accuracy, precision, and recall follow similar upward trends, surpassing 0.976, 0.976, 0.970, and 0.982, respectively. Log-loss decreases sharply to below 0.112, reflecting increasingly confident and well-calibrated predictions. The tight overlap between training and testing curves indicates minimal overfitting and strong generalization.

In the \textit{pocket milling}  scenario trained on 30,000 case studies (\hyperref[fig4_train_metrics]{\textcolor{blue}{Figure~\ref*{fig4_train_metrics}(b)}}), the model again converges quickly, with AUC rising above 0.984 early in training. F1-score, accuracy, precision, and recall surpass 0.957, 0.958, 0.957, and 0.958, demonstrating consistent gains in identifying both manufacturable and non-manufacturable designs. Log-loss steadily decreases to below 0.173, and the narrow train–test gap confirms reliable and stable learning.

The \textit{combined drilling-milling} scenario containing 40,000 case studies (\hyperref[fig4_train_metrics]{\textcolor{blue}{Figure~\ref*{fig4_train_metrics}(c)}}) represents a more challenging task due to the interaction between hole and pocket features. Nevertheless, the KAN achieves strong performance: AUC climbs quickly and stabilizes above 0.940, while F1-score, accuracy, precision, and recall exceed 0.898, 0.896, 0.870, and 0.928, respectively. Log-loss continues its downward trajectory, falling below 0.330, though remaining slightly higher than in single-feature scenarios, consistent with increased problem complexity. Training and testing curves remain closely aligned, indicating robust generalization even in higher-dimensional design spaces.

Overall, model performance decreases modestly as geometric complexity increases, yet remains consistently high across all scenarios. The KAN exhibits stable optimization behavior, smooth performance improvements, and minimal divergence between training and testing metrics, confirming its ability to learn meaningful and generalizable relationships between design parameters and manufacturability outcomes.

\begin{figure*}[!h]
    \centering
    \includegraphics[width=1\linewidth]{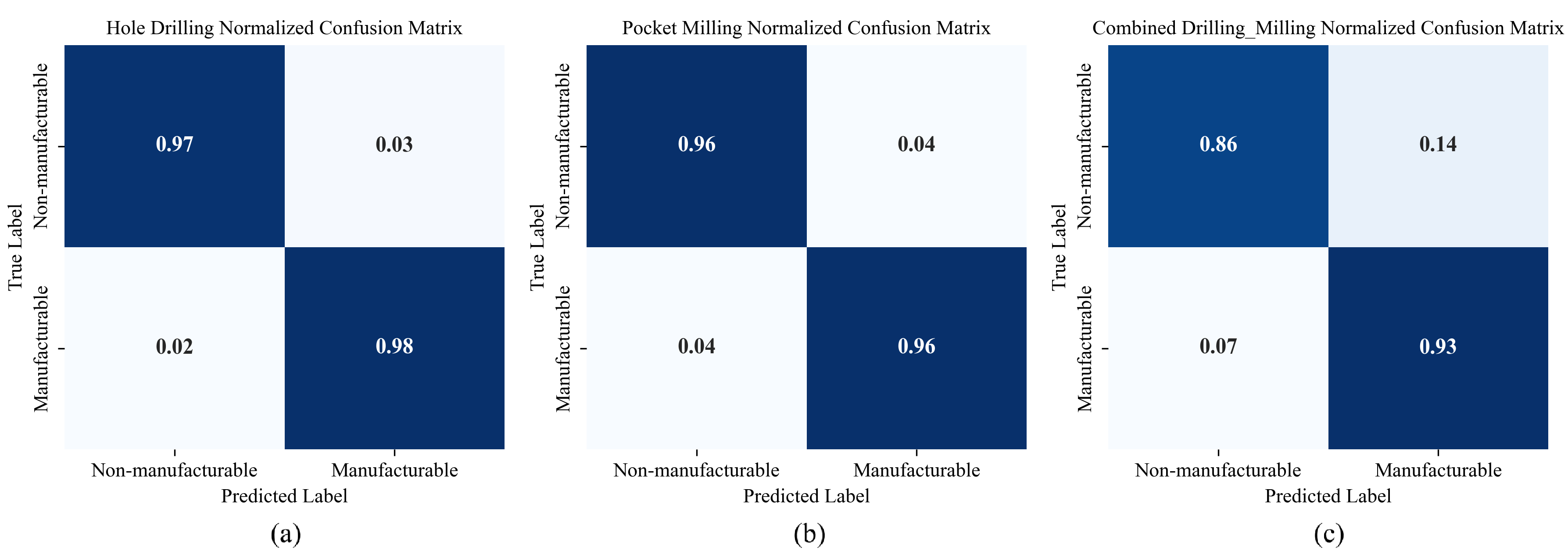}
    \caption{Normalized confusion matrices for the KAN model across: (a) Hole Drilling, (b) Pocket Milling, and (c) Combined Drilling–Milling scenarios.}
    \label{fig5_cmatrix}
\end{figure*}

\hyperref[fig5_cmatrix]{\textcolor{blue}{Figure~\ref*{fig5_cmatrix}}} present the normalized confusion matrices for all three manufacturing scenarios, summarizing the KAN model’s classification performance on the held-out test sets and demonstrating its ability to distinguish between manufacturable and non-manufacturable configurations consistently. In the \textit{hole drilling} scenario (\hyperref[fig5_cmatrix]{\textcolor{blue}{Figure~\ref*{fig5_cmatrix}(a)}}), the model achieves exceptionally high accuracy for both classes, correctly identifying 97\% of true-negative (TN), meaning non-manufacturable cases and 98\% of true-positive (TP), meaning manufacturable cases. The correspondingly low false-positive (FP) rate (3\%) and false-negative (FN) rate (2\%) indicate that the model reliably captures the geometric and tolerance-related constraints governing drilled-hole manufacturability.

For the \textit{pocket milling} scenario (\hyperref[fig5_cmatrix]{\textcolor{blue}{Figure~\ref*{fig5_cmatrix}(b)}}), the model shows similarly strong performance, correctly classifying 96\% of both non-manufacturable and manufacturable pockets. The slightly higher false-positive and false-negative rates (4\%) reflect the added complexity of pocket geometry compared to a hole due to more design parameters and tolerances, yet overall performance remains robust.

The \textit{combined drilling-milling} scenario (\hyperref[fig5_cmatrix]{\textcolor{blue}{Figure~\ref*{fig5_cmatrix}(c)}}) presents the most challenging task due to the interaction between hole and pocket features. Despite this increased complexity, the KAN maintains high discriminative power, correctly identifying 86\% of non-manufacturable designs and 93\% of manufacturable ones. This demonstrates the model’s ability to capture manufacturability behavior in multi-feature design spaces. Again, increases in false-positive (14\%) and false-negative (7\%) rates indicate the added complexity of feature interactions. The increased complexity of the decision boundaries and number of non-manufacturability conditions from 20 to 252 explained in \hyperref[sec4_3_data_generation]{\textcolor{blue}{Section~\ref*{sec4_3_data_generation}}} justifies the greater increase in the false-positive rate ($\sim10\%$) compared to the false-negative rate ($\sim3\%$) from single-feature scenarios to double-feature scenario.

Overall, the normalized confusion matrices confirm that the KAN model delivers highly reliable classification performance with balanced accuracy across both classes. These results demonstrate that the model effectively generalizes manufacturability constraints, from simple single-feature geometries to more complex multi-feature configurations, while maintaining low misclassification rates.

\subsection{Model Benchmarking Across Manufacturing Scenarios}
\label{sec6_3_benchmarking}

Having established that the KAN model achieves strong scenario-specific performance with stable convergence, high predictive accuracy, and consistent generalization across increasing levels of geometric complexity, the next step is to contextualize these results within the broader landscape of machine learning approaches. To this end, a comprehensive benchmarking study was conducted to compare the proposed KAN methodology against 14 widely used ML and DL models under identical training and evaluation conditions as described previously.

\rowcolors{2}{gray!10}{white}
\begin{table*}[t]
\centering
\small
\caption{Comparison of performance of various ML models across the Hole Drilling scenario, sorted by validation AUC score.}
\label{table7_drill_comparison}
\begin{tabular}{@{}clcccccc@{}}
\toprule
\textbf{Rank} & \textbf{Model} & \textbf{Val. AUC} & \textbf{Val. F1 score} & \textbf{Val. Accuracy} & \textbf{Val. Precision} & \textbf{Val. Recall} & \textbf{Val. Loss} \\ 
\midrule
1  & \textbf{KAN (Our model)} & \textbf{0.9919} & \textbf{0.9760} & \textbf{0.9760} & \textbf{0.9701} & \textbf{0.9818} & \textbf{0.1123} \\
2  & MLP             & 0.9831 & 0.9348 & 0.9343 & 0.9332 & 0.9364 & 0.1629 \\
3  & FNN             & 0.9792 & 0.9275 & 0.9259 & 0.9131 & 0.9424 & 0.1809 \\
4  & XGBoost         & 0.9790 & 0.9311 & 0.9289 & 0.9092 & 0.9540 & 0.1923 \\
5  & LightGBM        & 0.9790 & 0.9301 & 0.9274 & 0.9023 & 0.9597 & 0.1832 \\
6  & GBDT            & 0.9696 & 0.9108 & 0.9065 & 0.8753 & 0.9494 & 0.2359 \\
7  & ET              & 0.9663 & 0.9000 & 0.8919 & 0.8413 & 0.9676 & 0.2703 \\
8  & CatBoost        & 0.9658 & 0.9095 & 0.9056 & 0.8788 & 0.9424 & 0.2400 \\
9  & RF              & 0.9641 & 0.9013 & 0.8951 & 0.8552 & 0.9527 & 0.2832 \\
10 & SVM             & 0.9226 & 0.8535 & 0.8443 & 0.8100 & 0.9020 & 0.3429 \\
11 & DT              & 0.9079 & 0.8648 & 0.8540 & 0.8096 & 0.9280 & 0.3566 \\
12 & KNN             & 0.8881 & 0.8231 & 0.7969 & 0.7324 & 0.9394 & 0.4587 \\
13 & LR              & 0.8371 & 0.7911 & 0.7820 & 0.7635 & 0.8207 & 0.4947 \\
14 & RC              & 0.8323 & 0.7927 & 0.7784 & 0.7487 & 0.8421 & 0.5795 \\
15 & GNB             & 0.8009 & 0.7709 & 0.7468 & 0.7073 & 0.8471 & 0.5440 \\
\bottomrule
\end{tabular}
\end{table*}

\rowcolors{2}{gray!10}{white}
\begin{table*}[t]
\centering
\small
\caption{Comparison of performance of various ML models across the Pocket Milling scenario, sorted by validation AUC score.}
\label{table8_pocktet_comparison}
\begin{tabular}{@{}clcccccc@{}}
\toprule
\textbf{Rank} & \textbf{Model} & \textbf{Val. AUC} & \textbf{Val. F1 score} & \textbf{Val. Accuracy} & \textbf{Val. Precision} & \textbf{Val. Recall} & \textbf{Val. Loss} \\ 
\midrule
1  & \textbf{KAN (Our model)} & \textbf{0.9841} & \textbf{0.9575} & \textbf{0.9580} & \textbf{0.9573} & \textbf{0.9576} & \textbf{0.1727} \\
2  & FNN             & 0.9621 & 0.9040 & 0.9019 & 0.8744 & 0.9357 & 0.2349 \\
3  & XGBoost         & 0.9576 & 0.8985 & 0.8941 & 0.8532 & 0.9488 & 0.2674 \\
4  & LightGBM        & 0.9551 & 0.8935 & 0.8891 & 0.8495 & 0.9424 & 0.2600 \\
5  & CatBoost        & 0.9416 & 0.8763 & 0.8710 & 0.8322 & 0.9253 & 0.3070 \\
6  & GBDT            & 0.9397 & 0.8802 & 0.8743 & 0.8316 & 0.9347 & 0.3103 \\
7  & MLP             & 0.9389 & 0.8750 & 0.8712 & 0.8399 & 0.9131 & 0.3010 \\
8  & RF              & 0.9294 & 0.8694 & 0.8608 & 0.8101 & 0.9380 & 0.3649 \\
9  & ET              & 0.9259 & 0.8586 & 0.8446 & 0.7793 & 0.9559 & 0.3498 \\
10 & DT              & 0.8785 & 0.8365 & 0.8221 & 0.7658 & 0.9215 & 0.4070 \\
11 & SVM             & 0.8681 & 0.7982 & 0.7887 & 0.7553 & 0.8461 & 0.4416 \\
12 & LR              & 0.8172 & 0.7481 & 0.7446 & 0.7292 & 0.7680 & 0.5290 \\
13 & RC              & 0.8144 & 0.7500 & 0.7433 & 0.7224 & 0.7798 & 0.5927 \\
14 & KNN             & 0.8143 & 0.7644 & 0.7373 & 0.6860 & 0.8630 & 0.5513 \\
15 & GNB & 0.7562 & 0.7023 & 0.6978 & 0.6837 & 0.7219 & 0.5959 \\
\bottomrule
\end{tabular}
\end{table*}

\rowcolors{2}{gray!10}{white}
\begin{table*}[t]
\centering
\small
\caption{Comparison of performance of various ML models across the Combined Drilling–Milling scenario, sorted by validation AUC score.}
\label{table9_combined_comparison}
\begin{tabular}{@{}clcccccc@{}}
\toprule
\textbf{Rank} & \textbf{Model} & \textbf{Val. AUC} & \textbf{Val. F1 score} & \textbf{Val. Accuracy} & \textbf{Val. Precision} & \textbf{Val. Recall} & \textbf{Val. Loss} \\ 
\midrule
1  & \textbf{KAN (Our model)}  & \textbf{0.9406} & \textbf{0.8983} & \textbf{0.8961} & \textbf{0.8702} & \textbf{0.9282} & \textbf{0.3298} \\
2  & XGBoost          & 0.9120 & 0.8373 & 0.8306 & 0.7980 & 0.8806 & 0.3690 \\
3  & LightGBM         & 0.9082 & 0.8340 & 0.8270 & 0.7939 & 0.8783 & 0.3777 \\
4  & MLP              & 0.9025 & 0.8243 & 0.8226 & 0.8081 & 0.8412 & 0.3907 \\
5  & CatBoost         & 0.8938 & 0.8191 & 0.8104 & 0.7757 & 0.8677 & 0.4188 \\
6  & GBDT             & 0.8878 & 0.8148 & 0.8054 & 0.7702 & 0.8649 & 0.4273 \\
7  & RF               & 0.8671 & 0.7977 & 0.7835 & 0.7419 & 0.8626 & 0.4939 \\
8  & ET               & 0.8664 & 0.7918 & 0.7761 & 0.7332 & 0.8605 & 0.4874 \\
9  & FNN              & 0.8616 & 0.7894 & 0.7791 & 0.7471 & 0.8369 & 0.4482 \\
10 & SVM              & 0.8185 & 0.7549 & 0.7427 & 0.7140 & 0.8007 & 0.5147 \\
11 & DT               & 0.7885 & 0.7528 & 0.7138 & 0.6572 & 0.8809 & 0.5297 \\
12 & KNN              & 0.7741 & 0.7326 & 0.6752 & 0.6181 & 0.8992 & 0.6205 \\
13 & RC               & 0.7288 & 0.6814 & 0.6707 & 0.6536 & 0.7116 & 0.6334 \\
14 & LR               & 0.7278 & 0.6764 & 0.6682 & 0.6535 & 0.7009 & 0.6032 \\
15 & GNB              & 0.6899 & 0.6745 & 0.6255 & 0.5917 & 0.7842 & 0.6523 \\
\bottomrule
\end{tabular}
\end{table*}

\hyperref[table7_drill_comparison]{\textcolor{blue}{Table~\ref*{table7_drill_comparison}}}, \hyperref[table8_pocktet_comparison]{\textcolor{blue}{Table~\ref*{table8_pocktet_comparison}}}, and \hyperref[table9_combined_comparison]{\textcolor{blue}{Table~\ref*{table9_combined_comparison}}} summarize the benchmark validation metrics for the \textit{hole drilling}, \textit{pocket milling}, and \textit{combined drilling–milling} scenarios, respectively, with all models sorted based on validation AUC. Across all scenarios, the KAN model consistently achieved the highest performance, outperforming all other ML and DL baselines. Tree-based ensemble models such as XGBoost, LightGBM, GBDT, RF, and ET formed the next strongest group, exhibiting strong capability for capturing nonlinear feature interactions but still falling short of the discriminative power, stability, and generalization achieved by the KAN. Fully connected NNs, including MLP and FNN, delivered competitive results in simpler, single-feature scenarios but showed reduced performance as geometric complexity increased. Classical algorithms, such as SVM, LR, RC, KNN, and GNB, consistently ranked lower, reflecting their limited ability to model the nonlinear, geometry- and tolerance-dependent relationships inherent in manufacturability prediction. The superior and consistent performance of the KAN model underscores the importance of architectures specifically designed to learn smooth, interpretable functional relationships between design parameters and manufacturability constraints.

Unlike dense neural networks or tree-based models, which approximate the feature space through either weighted linear transformations or discontinuous partitioning, KANs leverage spline-based functional approximations that naturally capture the continuous, nonlinear dependencies embedded in geometric design rules and tolerance interactions. This advantage becomes even more pronounced as the design space grows in dimensionality and complexity, explaining the model’s sustained top performance in the \textit{combined drilling–milling} scenario where multiple features interact simultaneously. Overall, the benchmarking analysis confirms that the KAN architecture provides the most accurate, robust, and generalizable manufacturability predictions, highlighting its suitability as a high-fidelity and interpretable solution for parameter-based engineering design evaluation.

\subsection{Interpretability Analysis}
\label{sec6_4_interpretability}

\begin{figure*}[h]
    \centering
    \includegraphics[width=1\linewidth]{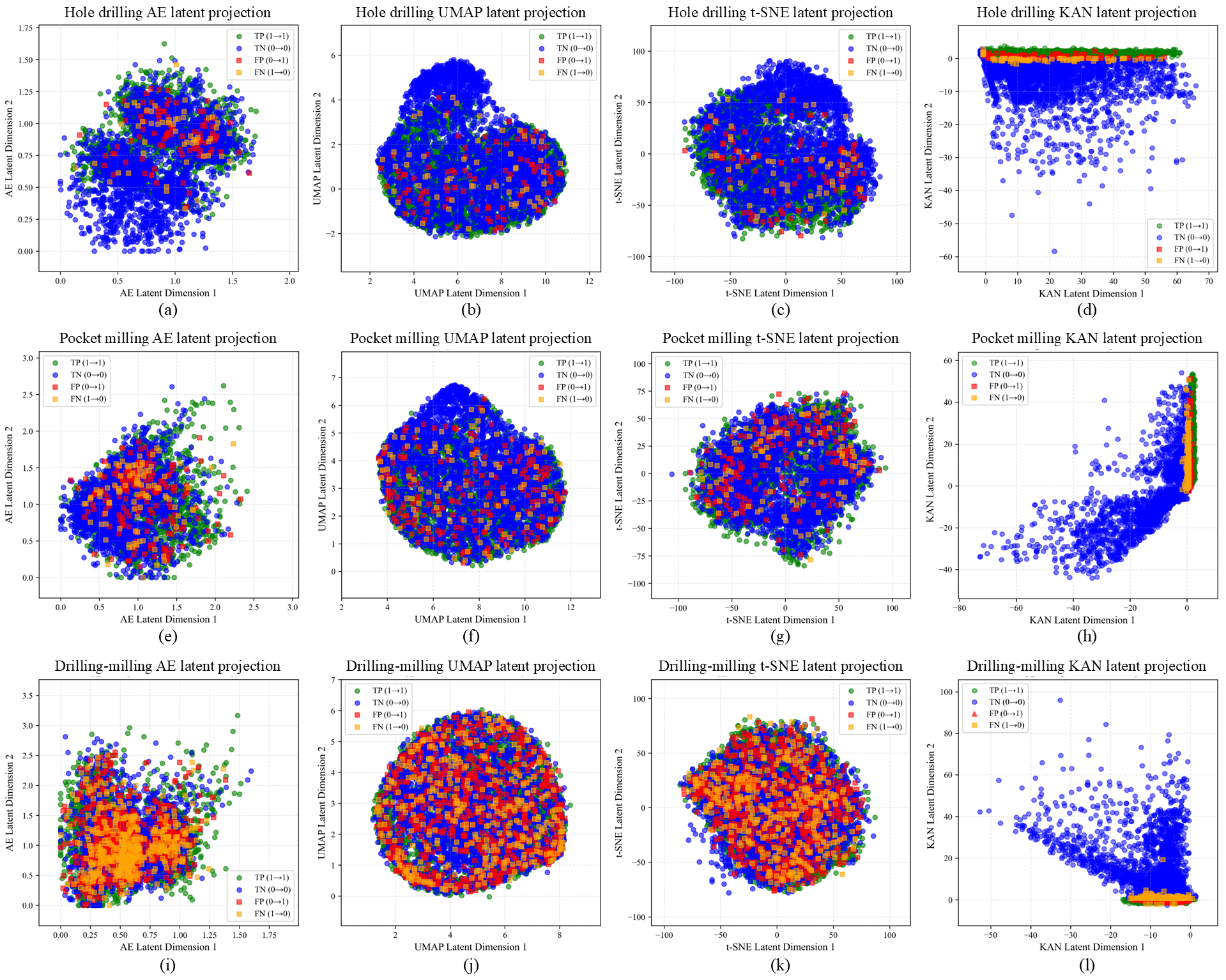}
    \caption{Latent space projections of the hole drilling (a–d), pocket milling (e–h), and combined drilling–milling (i–l).}
    \label{fig7_KAN_latent}
\end{figure*}

To further investigate the internal structure of the design space and assess whether low-dimensional embeddings can reveal meaningful patterns related to manufacturability, three widely used dimensionality reduction and embedding techniques, including AE with a two-neuron bottleneck, UMAP, and t-SNE, were employed as described in \hyperref[sec5_5_interpret_tools]{\textcolor{blue}{Section~\ref*{sec5_5_interpret_tools}}}. \hyperref[fig7_KAN_latent]{\textcolor{blue}{Figure~\ref*{fig7_KAN_latent}}} illustrates the resulting two-dimensional latent projections, where \hyperref[fig7_KAN_latent]{\textcolor{blue}{Figure~\ref*{fig7_KAN_latent}(a-c)}}, \hyperref[fig7_KAN_latent]{\textcolor{blue}{Figure~\ref*{fig7_KAN_latent}(e-g)}}, and \hyperref[fig7_KAN_latent]{\textcolor{blue}{Figure~\ref*{fig7_KAN_latent}(i-k)}} correspond to the \textit{hole drilling}, \textit{pocket milling}, and \textit{combined drilling-milling} scenarios, respectively.

Across all three methods, the embedded manifolds remain highly entangled, with no meaningful separation between manufacturable and non-manufacturable cases. TP, TN, FP, and FN predictions appear uniformly intermixed, forming dense, overlapping clusters without discernible boundaries or geometric structure. This behavior reflects the inherent complexity of manufacturability assessment, particularly when tolerance interactions are included, where slight variations in geometric dimensions or clearance margins can induce abrupt changes in manufacturability status. Because these interactions produce highly complex and nonlinear decision regions, unsupervised or reconstruction-based embeddings are unable to extract the underlying discriminative relationships needed to distinguish outcomes. Consequently, AE, UMAP, and t-SNE fail to provide actionable insight or interpretable clustering for this problem, underscoring the limitations of conventional manifold-learning techniques in capturing the intricate constraints governing manufacturability.

To address these limitations, the latent space learned internally by the KAN model was directly visualized by constraining its final hidden layer to two neurons and projecting their outputs into two dimensions. \hyperref[fig7_KAN_latent]{\textcolor{blue}{Figure~\ref*{fig7_KAN_latent}(d), (h), and (l)}} depict the resulting KAN latent spaces for for \textit{hole drilling}, \textit{pocket milling}, and \textit{combined drilling-milling} respectively. Unlike the unsupervised embeddings, the KAN projections exhibit a clear, structured organization that aligns with the underlying logic of manufacturability. In the \textit{hole drilling} scenario, TN predictions occupy a broad and continuous region, while TP predictions form a compact band near the top of the manifold, with misclassifications confined to narrow transitional zones. The \textit{pocket milling} scenario displays a similarly coherent progression, reflecting the complex relationships among pocket depth, internal radius, tool constraints, and tolerance thresholds. Even in the \textit{combined drilling–milling} case, the most geometrically complex, the KAN learns a smooth, physically interpretable trajectory along which manufacturable and non-manufacturable designs separate naturally, with FP and FN concentrated near regions corresponding to borderline geometric conditions.

These visualizations demonstrate that the KAN does more than merely classify, and it constructs a meaningful, low-dimensional coordinate system that compactly encodes manufacturability behavior. The resulting latent manifolds offer interpretable, geometry-aware insight into how different design configurations relate to manufacturability constraints. The contrast between the KAN latent spaces and the conventional AE, UMAP, and t-SNE embeddings further highlights a central advantage of the proposed methodology. By learning functional, spline-based transformations from design parameters to manufacturability outcomes, the KAN develops inherently structured representations grounded in the physics and logic of design rules. This structured latent space not only separates manufacturable from non-manufacturable regions more coherently but also provides a principled foundation for deeper interpretability.

\begin{figure*}[h]
    \centering
    \includegraphics[width=1\linewidth]{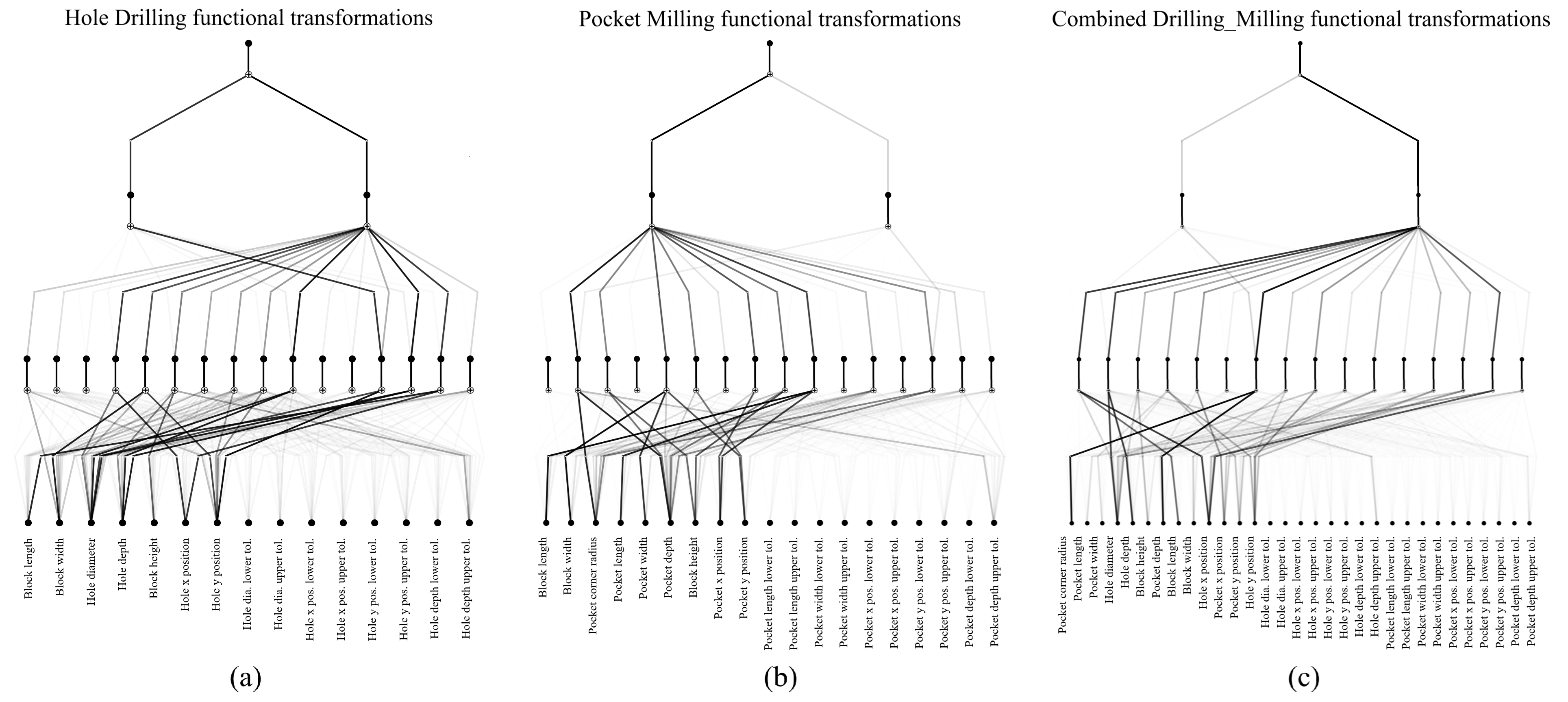}
    \caption{Learned spline-based functional transformations within the KAN for (a) Hole Drilling, (b) Pocket Milling, and (c) Combined Drilling–Milling.}
    \label{fig8_KAN_splines}
\end{figure*}

Building upon the latent-space analysis presented above, we now examine how the KAN model learns and represents the nonlinear relationships between geometric parameters, tolerance values, and manufacturability outcomes. Unlike conventional NNs, which distribute learned behavior across dense weight matrices, KANs provide explicit functional transparency through their univariate spline transformations. Each edge in the KAN architecture corresponds to a learned spline function $\phi_{}(x)$ that maps an input parameter (or the output of a previous layer) to a transformed representation contributing to the final prediction. Visualizing these functions reveals, in a direct and interpretable manner, how the model internally processes each design feature.

\hyperref[fig8_KAN_splines]{\textcolor{blue}{Figure~\ref*{fig8_KAN_splines}}} illustrates the learned spline functions across the three scenarios mentioned above. While these figures aggregate many splines and do not isolate each individually, the overall shape diversity, intensity, and variability reveal how the KAN adapts its functional basis to different levels of geometric and tolerance complexity. In the \textit{hole drilling} scenario (\hyperref[fig8_KAN_splines]{\textcolor{blue}{Figure~\ref*{fig8_KAN_splines}(a)}}), splines associated with hole diameter, depth, block height, and their respective tolerance bounds show the most substantial variations, reflecting the central role of diameter–depth feasibility and remaining-bottom-thickness constraints in manufacturability. Positional parameters and their tolerances also contribute significantly by enforcing minimum wall thickness requirements. Among tolerance features, the upper bound of hole depth emerges as the most influential. Despite the inclusion of tolerance-driven micro-variations, these splines remain relatively coherent, consistent with the structured and rule-driven nature of drilling manufacturability.

In the \textit{pocket milling} scenario (\hyperref[fig8_KAN_splines]{\textcolor{blue}{Figure~\ref*{fig8_KAN_splines}(b)}}), the spline set becomes more diverse. Pocket depth, internal corner radius, block height, and their tolerance bounds exhibit pronounced nonlinearities, corresponding to tool-diameter limitations, allowable depth, and bottom-thickness requirements. Positional parameters again play a significant role by governing wall-thickness constraints. As in drilling, the upper depth tolerance is the most influential parameter among tolerances.

The \textit{combined drilling-milling} scenario (\hyperref[fig8_KAN_splines]{\textcolor{blue}{Figure~\ref*{fig8_KAN_splines}(c)}}) exhibits the greatest diversity and dispersion of spline functions. Here, manufacturability is governed by simultaneous cross-feature interactions between the hole and the pocket. Spline functions linked to pocket depth, corner radius, hole diameter, and hole depth contribute most strongly, with positional parameters capturing secondary interactions. Overall block and pocket dimensions play a tertiary but still visible role. The broader spread of spline behaviors indicates that the KAN must learn a richer set of nonlinear transformations to accommodate these intertwined constraints.

Collectively, these spline visualizations do not serve as per-feature diagnostic tools but instead reveal how the KAN internally organizes and transforms parametric information. The clear progression, from relatively coherent splines in drilling to moderate variability in pocket milling to highly diverse splines in the combined scenario, directly mirrors the escalating complexity of the underlying design space. To identify which specific parameters most strongly drive manufacturability decisions and to provide actionable, feature-level guidance to designers, we next employ SHAP analysis.

\begin{figure*}[h]
    \centering
    \includegraphics[width=1\linewidth]{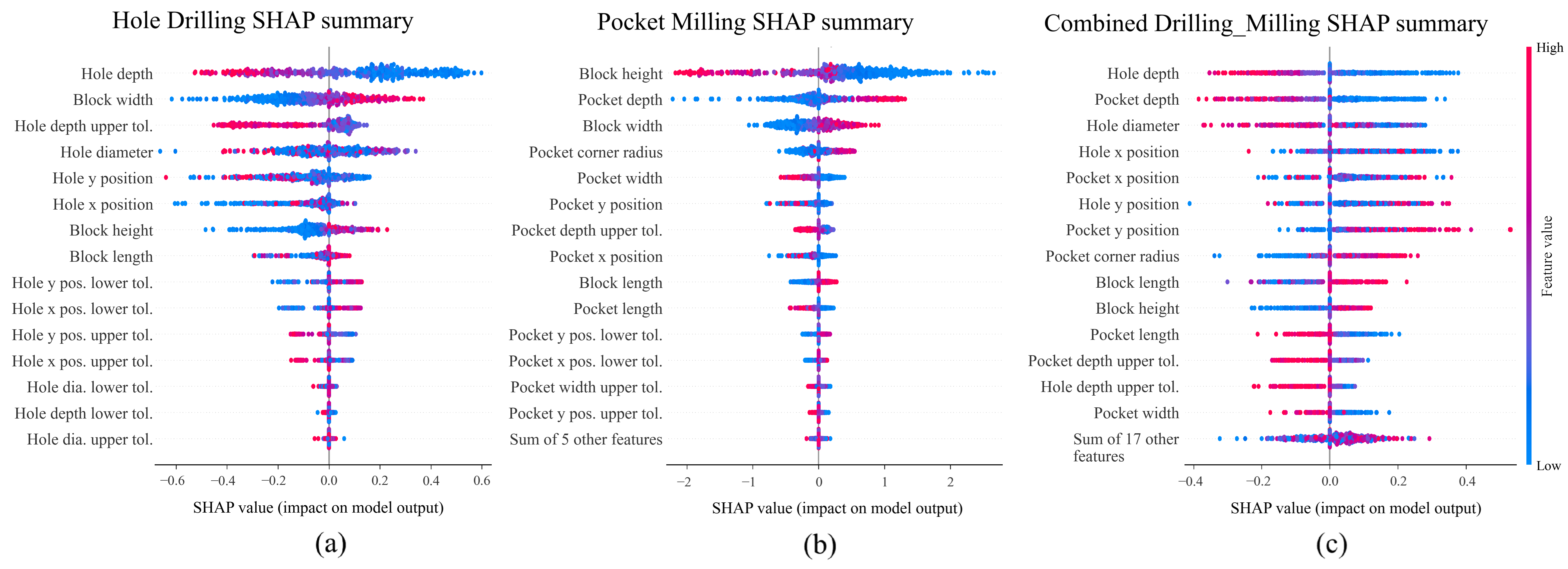}
    \caption{SHAP summary plots for (a) Hole Drilling, (b) Pocket Milling, and (c) Combined Drilling–Milling.}
    \label{fig9_shap_summary}
\end{figure*}

The SHAP summary plots illustrated in \hyperref[fig9_shap_summary]{\textcolor{blue}{Figure~\ref*{fig9_shap_summary}}} quantify how individual geometric- and tolerance-related parameters influence the KAN’s manufacturability predictions and complement the functional patterns observed in the spline visualizations previously discussed. In the \textit{hole drilling} scenario (\hyperref[fig9_shap_summary]{\textcolor{blue}{Figure~\ref*{fig9_shap_summary}(a)}}), SHAP values highlight hole depth, block width, upper depth tolerance, hole diameter, and hole position as the most influential features. Increasing hole depth or its upper tolerance limit shifts predictions toward non-manufacturability by violating depth-to-diameter and bottom-thickness constraints, whereas increasing hole diameter or block width generally improves manufacturability. Hole position also plays a strong role, as moving the hole closer to block edges reduces manufacturability by reducing wall thickness.

In the \textit{pocket milling} scenario (\hyperref[fig9_shap_summary]{\textcolor{blue}{Figure~\ref*{fig9_shap_summary}(b)}}), pocket depth, corner radius, block height, block width, and positional tolerances have the most significant influence. Larger pocket depths consistently decrease manufacturability because they require longer tools and risk insufficient bottom thickness, while increasing the corner radius improves manufacturability by enabling larger cutters and reducing cutting loads. Block dimensions again have a positive effect, whereas pocket position and its upper depth tolerance reduce manufacturability when pockets approach geometric limits or violate wall-thickness constraints.

The \textit{combined drilling-milling} scenario (\hyperref[fig9_shap_summary]{\textcolor{blue}{Figure~\ref*{fig9_shap_summary}(c)}}) exhibits the highest complexity, with both hole- and pocket-related depth parameters showing strong negative contributions; increasing either feature’s depth or upper tolerance limit pushes the design toward non-manufacturability due to tool-access and bottom-thickness constraints. Conversely, increases in hole diameter and corner radius generally enhance manufacturability, whereas positional offsets often decrease it when interactions between features result in thin or interfering walls. Across all three scenarios, the SHAP results provide clear, directional insight into how specific parameters influence manufacturability and validate the KAN’s ability to learn and reflect established DFM principles.

Across all evaluation components, including training dynamics, confusion matrix performance, benchmarking against alternative models, latent space visualization, KAN spline analyses, and SHAP-based feature attribution, the results demonstrate that the proposed KAN-based approach provides a highly reliable, interpretable, and computationally efficient framework for manufacturability assessment. The model consistently achieves strong predictive accuracy across the explained scenarios, generalizing well even under increased geometric complexity and tolerance-induced variability. Interpretability analyses reveal that the KAN not only captures the fundamental DFM rules governing each process but also quantifies how design parameters and their tolerances influence manufacturability outcomes. Taken together, these results confirm that the proposed tabular, parameter-based methodology, powered by KAN’s functional transparency, offers a robust and scalable foundation for data-driven manufacturability evaluation across diverse mechanical design tasks.

\section{Case Study}
\label{sec7_CaseStudy}

To experimentally validate the proposed methodology and demonstrate its practical application in real engineering scenarios, a representative case study involving both a hole and a pocket feature was designed, evaluated, iteratively modified, and ultimately manufactured. The case was intentionally constructed to include multiple forms of non-manufacturability inspired by common design mistakes made by novice designers. Specifically, the initial design incorporates at least one violation from each major manufacturability constraint category, including process-related infeasibility, component structural integrity limitations, and tooling-related restrictions, as previously explained in \hyperref[sec4_3_data_generation]{\textcolor{blue}{Section~\ref*{sec4_3_data_generation}}}.

The manufacturability of the initial design is assessed through an iterative, progressive refinement process. The design is first analyzed using the proposed KAN-based model. If predicted to be non-manufacturable, feature-level modification guidance is extracted through case-specific interpretability analyses, and the necessary parameter adjustments are applied to generate an improved design. The updated model is then re-evaluated, and this cycle continues until a fully manufacturable configuration is achieved. The final modified design is fabricated to validate the correctness and reliability of the proposed approach. Additionally, to further illustrate the complexity of real-world challenges, an intermediate modified design is also manufactured to demonstrate the types of geometric defects and process failures that the model successfully identified and prevented.

\begin{figure}[t]
    \centering
    \includegraphics[width=1\linewidth]{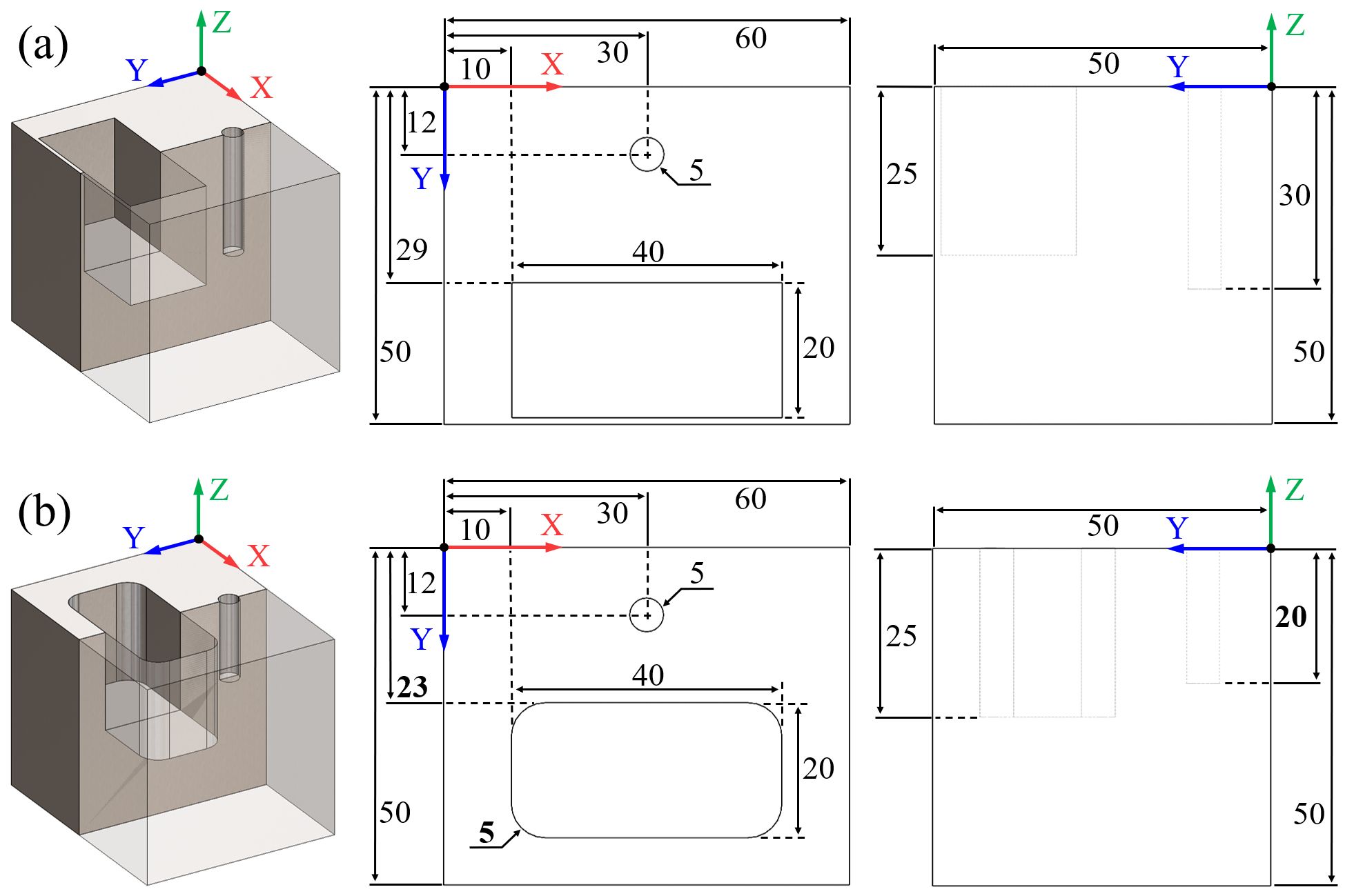}
    \caption{Case study dimensions: (a) initial non-manufacturable configuration, and (b) final manufacturable version after iterative modification.}
    \label{fig10_Casestudy_geometries}
\end{figure}

\hyperref[fig10_Casestudy_geometries]{\textcolor{blue}{Figure~\ref*{fig10_Casestudy_geometries}}}  presents the geometric definitions and dimensional specifications of the case study. \hyperref[fig10_Casestudy_geometries]{\textcolor{blue}{Figure~\ref*{fig10_Casestudy_geometries}(a)}} shows the initial non-manufacturable design, which includes multiple geometric and tooling-related violations such as sharp internal corners, insufficient wall thickness, and an overly deep hole. \hyperref[fig10_Casestudy_geometries]{\textcolor{blue}{Figure~\ref*{fig10_Casestudy_geometries}(b)}} illustrates the final modified design obtained after iterative correction guided by the proposed KAN-based manufacturability assessment framework. All dimensions are expressed in millimeters, and a symmetric tolerance of $\pm 0.1$ mm is applied to all hole- and pocket-related parameters. However, the tolerance values are omitted in this figure for clarity.

\begin{figure*}[h]
    \centering    \includegraphics[width=1\linewidth]{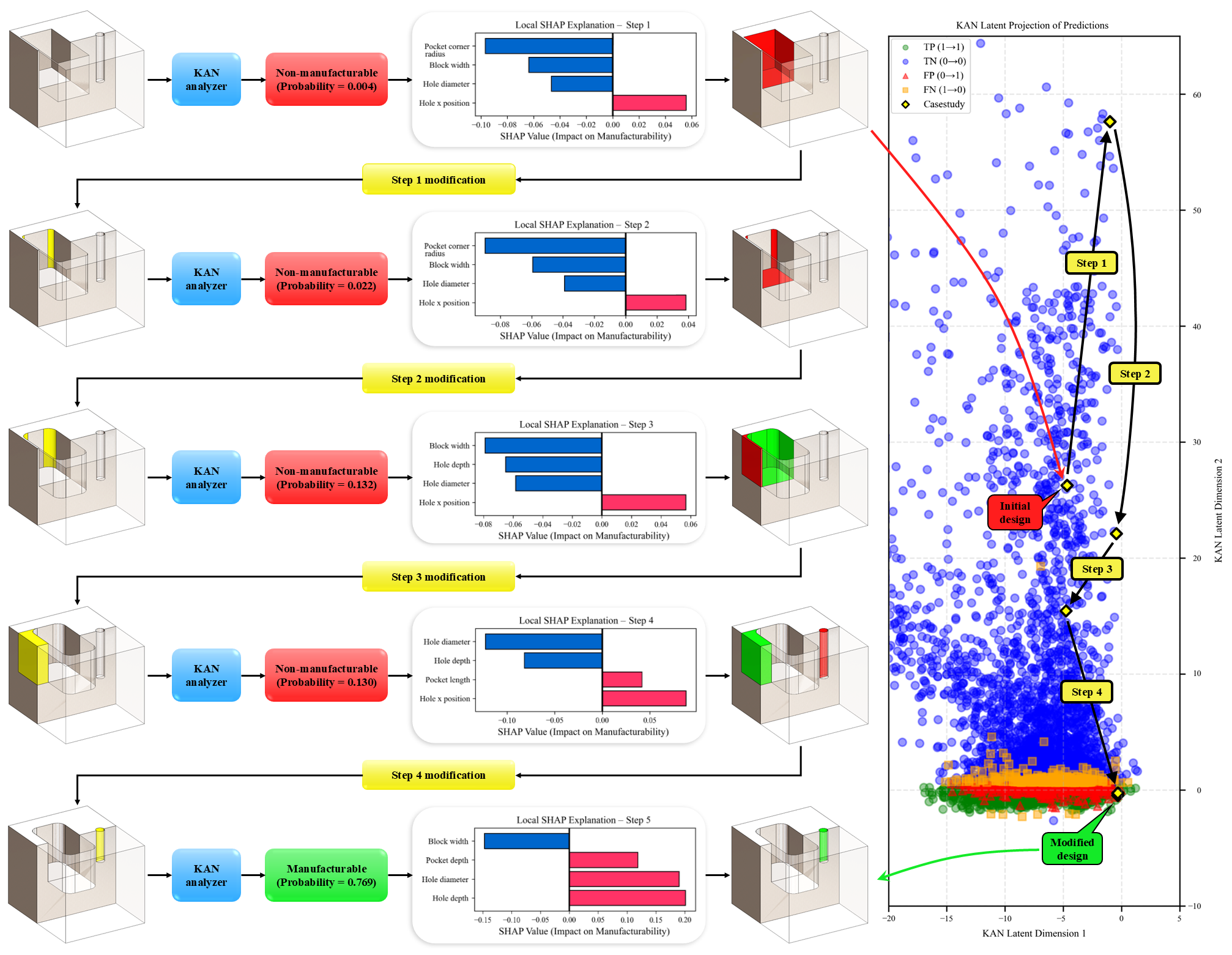}
    \caption{Iterative manufacturability evaluation and modification process for a representative case study using the proposed KAN-based approach.}
    \label{fig11_casestudy_modification_loop}
\end{figure*}

\hyperref[fig11_casestudy_modification_loop]{\textcolor{blue}{Figure~\ref*{fig11_casestudy_modification_loop}}} illustrates the four-step progressive evaluation–modification cycle applied to the initial design. In each iteration, the proposed KAN model predicts the manufacturability and provides case-specific interpretability insights highlighting which parameters contribute positively (red bars) or negatively (blue bars) to manufacturability. Parameters with negative contributions shift the design toward the non-manufacturable region of the decision space, whereas parameters with positive contributions support manufacturability. These insights guide targeted geometric modifications until a feasible design is achieved.

The iterative modification process begins with evaluating the initial design, where the model predicts a manufacturability probability of 0.004, indicating a clearly non-manufacturable configuration. The case-specific interpretability analysis shows that the most substantial negative contribution arises from the \textit{pocket corner radius}, correctly highlighting the presence of sharp internal corners that cannot be produced using a rotating end mill. Such corners violate basic milling feasibility rules, confirming that the initial design contains a fundamental geometric infeasibility.

To address this issue, the first modification introduces a 2.5 mm fillet to all internal pocket corners, enabling machining with a $\phi5$ mm cutter. After this adjustment, the updated design is re-evaluated, yielding a slightly higher manufacturability probability of 0.022, which remains non-manufacturable and shows only a slight improvement. The interpretability analysis again identifies the \textit{pocket corner radius} as the dominant negative factor, albeit for a different reason. Although the corners are no longer sharp, the tool diameter-to-pocket depth ratio is still violated, preventing cutter reachability at the required depth.

In the second modification, the \textit{pocket corner radius} is increased to 5 mm to satisfy the tool diameter-to-depth constraint. A new evaluation yields a manufacturability probability of 0.132, an improvement but still below the threshold. This time, the interpretability results show that the negative contribution originates from the \textit{block width} parameter, which interacts with \textit{pocket width} and \textit{pocket Y-location} to indicate an insufficient wall thickness. Because \textit{block width} is restricted by raw-stock dimensions and \textit{pocket width} is a required design constraint, the only feasible adjustment is to move the pocket. The model correctly identifies this multi-parameter interaction as the next cause of infeasibility.

The third modification shifts the pocket toward the positive Y direction, increasing the wall thickness from 1 mm to 7 mm. Evaluating this configuration gives a manufacturability probability of 0.130, still indicating a non-manufacturable design. The interpretability analysis now shows the combined influence of \textit{hole diameter} and \textit{hole depth}, revealing a violation of the drilling diameter-to-depth ratio constraint. Since \textit{hole diameter} is a required design parameter and cannot be changed, reducing \textit{hole depth} is the only feasible correction.

\begin{figure*}[h]
    \centering    \includegraphics[width=1\linewidth]{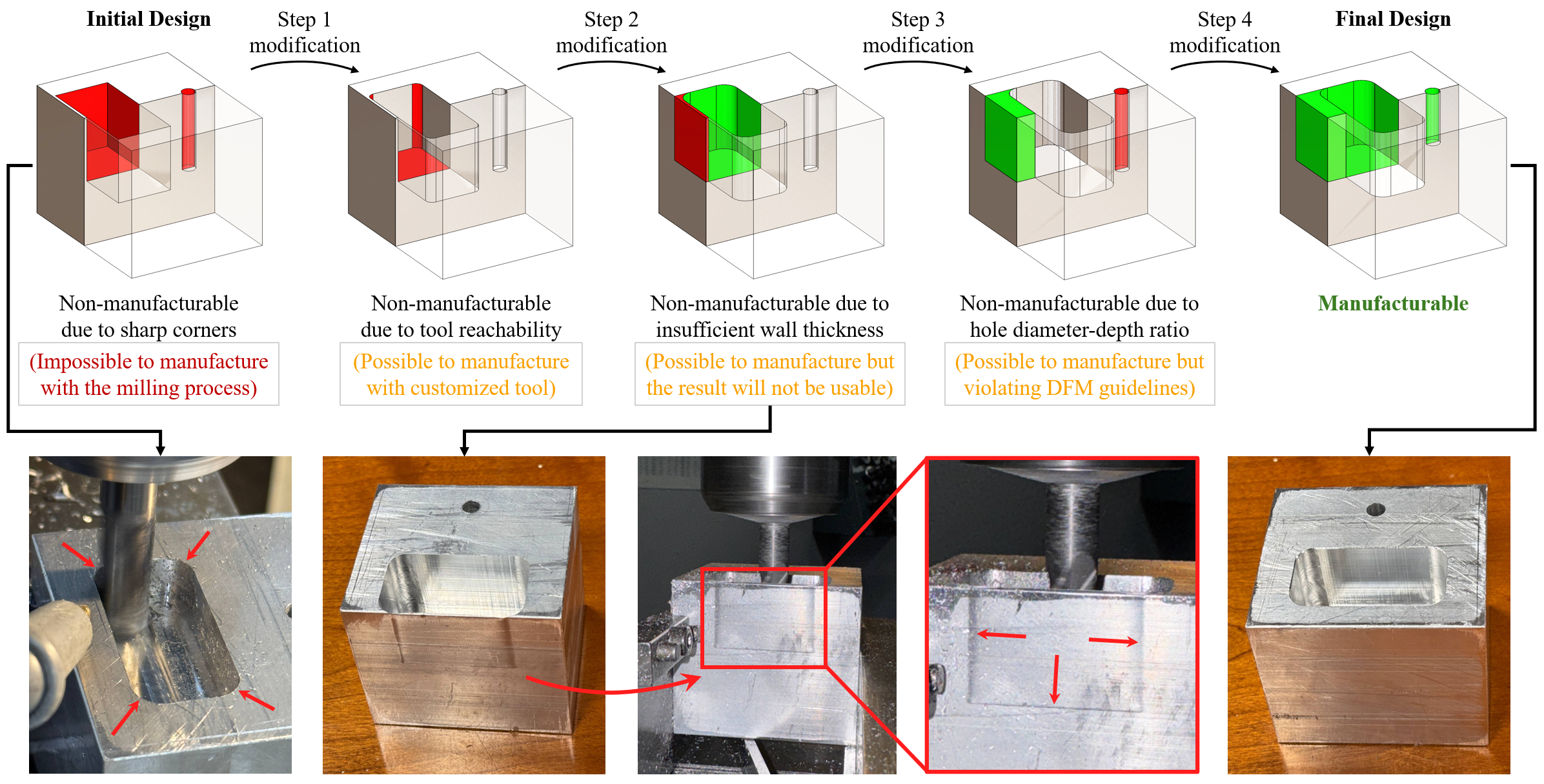}
    \caption{Experimental validation of the proposed KAN methodology outcomes and identified manufacturability violations, their physical representations in fabricated samples, and the final defect-free manufacturable design.}
    \label{fig12_casestudy_fabrication}
\end{figure*}

In the fourth modification, the \textit{hole depth} is reduced from 30 mm to 20 mm, resolving the dimensional ratio violation. The subsequent evaluation yields a manufacturability probability of 0.769, confirming that the design has transitioned into the manufacturable region. Through a sequence of four targeted modifications, each guided by case-specific interpretability feedback, the initial non-manufacturable design is progressively refined into a fully manufacturable configuration.

\hyperref[fig12_casestudy_fabrication]{\textcolor{blue}{Figure~\ref*{fig12_casestudy_fabrication}}} summarizes the manufacturability issues identified by the proposed model at each iteration and illustrates their real-world implications through fabricated components. At each modification stage, the model not only predicts manufacturability but also correctly isolates the parameters and parameter interactions responsible for non-manufacturability, even when those conditions arise from different categories of constraints.

In the initial design, the model accurately identifies the presence of a pocket with sharp internal corners, which cannot be produced with a rotational milling cutter. Although such features may be fabricated using electron discharge machining (EDM), the model is explicitly trained for milling and turning operations and correctly flags this geometry as non-manufacturable. This demonstrates the method’s ability to detect process-specific infeasibility.

After the first modification step, the pocket remains non-manufacturable due to tool reachability limitations. While specialized tooling could potentially machine this geometry, the prediction is consistent with the commercial tool availability embedded in the dataset, demonstrating the model’s capability to detect tooling-related constraints.

Following the second modification, the pocket becomes manufacturable with standard tools, but the overall design is still non-manufacturable due to insufficient wall thickness. This highlights the model’s ability to capture structural integrity constraints. To illustrate the real-world consequences of this violation, a physical component was fabricated at this intermediate stage. As shown in \hyperref[fig12_casestudy_fabrication]{\textcolor{blue}{Figure~\ref*{fig12_casestudy_fabrication}}}, the resulting part exhibits severe geometric distortion, rendering it defective and unusable.

After the third modification step, the design remains non-manufacturable, this time due to excessive hole depth relative to the allowable diameter-to-depth ratio defined by DFM guidelines. Although drilling a $\phi5$ mm hole to 30 mm depth is technically possible, it violates recommended machining practice, which the model correctly identifies as a manufacturability risk driven by DFM-informed constraints.

Finally, after reducing the hole depth to satisfy these DFM rules, the design becomes manufacturable and is successfully fabricated. The final component demonstrates full compliance with machining accessibility, tooling availability, structural integrity requirements, and DFM principles. Collectively, these results confirm that the model’s predictions are technically accurate, practically meaningful, and capable of guiding designers through reliable, actionable corrections of non-manufacturable designs in real engineering contexts.

\section{Conclusion}
\label{sec8_discussion}

This study introduced a new manufacturability assessment methodology that departs from geometry-dependent, CAD-driven pipelines and instead leverages a parametric, tolerance-aware, and functionally interpretable learning paradigm based on KANs. By operating directly on structured design parameters, the proposed approach eliminates the need for computationally intensive geometric preprocessing, prevents information loss due to CAD discretization, and enables seamless integration of dimensional tolerances that influence real-world manufacturability.

A large-scale dataset containing 300,000 labeled designs was developed spanning three representative machining scenarios, \textit{hole drilling}, \textit{pocket milling}, and \textit{combined drilling-milling}, integrating DFM rules, machining constraints, and commercially available tooling limitations. Comprehensive benchmarking across fourteen ML and DL models demonstrated that the KAN consistently delivered the highest accuracy, stability, and generalization, achieving validation AUC values of 0.9919, 0.9841, and 0.9406 across the three scenarios, respectively. Beyond predictive performance, the model provided a high level of interpretability through spline-function analysis, SHAP-based parameter attribution, and KAN-derived latent-space projections, offering both global and local insights into how geometric and tolerance parameters influence manufacturability outcomes.

An industrial case study further validated the practical utility of the proposed framework. Through iterative, parameter-level modifications informed by case-specific interpretability analysis, a highly non-manufacturable design was progressively transformed into a fully manufacturable component. The model's predicted structural and tooling-related defects were observed during fabrication of an intermediate design, while the final design was successfully produced, confirming that the model not only identifies manufacturability issues but also provides actionable, technically accurate guidance for resolving them.

Overall, the proposed KAN-based methodology constitutes a lightweight, transparent, and scalable alternative to geometry-driven DL approaches. It provides rapid manufacturability assessment, integrates tolerance effects, supports reversible design feedback, and operates in complete alignment with engineering semantics. This positions the method as a practical decision-support tool for designers, particularly in early design stages where rapid feedback and interpretability are essential.

Future work will extend this methodology to additional manufacturing processes, incorporate multi-objective criteria such as cost and time, and explore hybrid representations that integrate parametric and geometric information when needed. Expanding the interpretability capabilities and validating the framework across industrial-scale datasets will further enhance its robustness and broaden its applicability within modern Industry 5.0 design–manufacturing ecosystems.

\section*{CRediT authorship contribution statement}

\textbf{Masoud Deylami:} Writing – review \& editing, Writing – original draft, Visualization, Validation, Software, Methodology, Investigation, Formal analysis, Data curation, Conceptualization.
\textbf{Negar Izadipour:} Writing – review \& editing, Writing – original draft, Visualization, Software, Data curation. 
\textbf{Adel Alaeddini:} Supervision, Resources, Project administration, Funding acquisition.

\section*{Declaration of competing interest}

The authors declare that they have no known competing financial interests or personal relationships that could have appeared to influence the work reported in this paper.


\begin{thebibliography}{00}

\bibitem{yan2024}
Yan X, Williams R, Arvanitis E, Melkote S. Deep learning-based semantic segmentation of machinable volumes for cyber manufacturing service. \textit{J Manuf Syst} 2024;72:16–25. doi:\href{https://doi.org/10.1016/j.jmsy.2023.11.005}{10.1016/j.jmsy.2023.11.005}.


\bibitem{chen2025manu}
Chen Y, Cagan J, Kara LB. VIRL: Volume-informed representation learning towards few-shot manufacturability estimation. \textit{J Intell Manuf} 2025. doi:\href{https://doi.org/10.1007/s10845-025-02575-8}{10.1007/s10845-025-02575-8}.




\bibitem{doellken2020}
Doellken M, Zimmerer C, Matthiesen S. Challenges faced by design engineers when considering manufacturing in design – an interview study. \textit{Proc Des Soc Des Conf} 2020;1:837–46. doi:\href{https://doi.org/10.1017/dsd.2020.302}{10.1017/dsd.2020.302}.

\bibitem{jing2026}
Jing Y, Zhou G, Zhang C, Chang F. Human-centric proactive design for manufacturing with deep generative modeling in Industry 5.0. \textit{Adv Eng Inform} 2026;69(Pt B):103952. doi:\href{https://doi.org/10.1016/j.aei.2025.103952}{10.1016/j.aei.2025.103952}.

\bibitem{jarosz2024}
Jarosz K, Matijas T, Liu R. Investigating the human capabilities for intuitive evaluation of component manufacturability. \textit{Manuf Lett} 2024;41(Suppl):1618–23. doi:\href{https://doi.org/10.1016/j.mfglet.2024.09.188}{10.1016/j.mfglet.2024.09.188}.

\bibitem{abdelall2020}
Abdelall ES, Alawneh L, Eldakroury M. A manufacturability-based assessment and design modification support tool. \textit{J Appl Res Technol} 2020;18(6):410–24. doi:\href{https://doi.org/10.22201/icat.24486736e.2020.18.6.1366}{10.22201/icat.24486736e.2020.18.6.1366}.

\bibitem{ordek2025}
Ördek B, Borgianni Y. Experts’ and novices’ views on the use of additive manufacturing for the fabrication of parts based on their geometry. \textit{Int J Interact Des Manuf} 2025. doi:\href{https://doi.org/10.1007/s12008-025-02394-4}{10.1007/s12008-025-02394-4}.





\bibitem{xu2021}
Xu X, Lu Y, Vogel-Heuser B, Wang L. Industry 4.0 and Industry 5.0—Inception, conception and perception. \textit{J Manuf Syst} 2021;61:530–35. doi:\href{https://doi.org/10.1016/j.jmsy.2021.10.006}{10.1016/j.jmsy.2021.10.006}.

\bibitem{leng2023}
Leng J, Zhong Y, Lin Z, Xu K, Mourtzis D, Zhou X, Zheng P, Liu Q, Zhao JL, Shen W. Towards resilience in Industry 5.0: A decentralized autonomous manufacturing paradigm. \textit{J Manuf Syst} 2023;71:95–114. doi:\href{https://doi.org/10.1016/j.jmsy.2023.08.023}{10.1016/j.jmsy.2023.08.023}.

\bibitem{leng2022}
Leng J, Sha W, Wang B, Zheng P, Zhuang C, Liu Q, Wuest T, Mourtzis D, Wang L. Industry 5.0: Prospect and retrospect. \textit{J Manuf Syst} 2022;65:279–95. doi:\href{https://doi.org/10.1016/j.jmsy.2022.09.017}{10.1016/j.jmsy.2022.09.017}.

\bibitem{zhang2023}
Zhang C, Wang Z, Zhou G, Chang F, Ma D, Jing Y, Cheng W, Ding K, Zhao D. Towards new-generation human-centric smart manufacturing in Industry 5.0: A systematic review. \textit{Adv Eng Inform} 2023;57:102121. doi:\href{https://doi.org/10.1016/j.aei.2023.102121}{10.1016/j.aei.2023.102121}.




\bibitem{chen2025}
Chen JC, Kakati A. Development of e-Dart-based artificial neural network for multiple quality characteristic online defect detection system in injection molding processes. \textit{Int J Adv Manuf Technol} 2025;136:951–60. doi:\href{https://doi.org/10.1007/s00170-024-14872-2}{10.1007/s00170-024-14872-2}.

\bibitem{diogo2025}
Diogo T, Ramos A, Pereira F, et al. Automated detection and classification of soldering defects in printed circuit boards using deep learning and optical and thermal imaging. \textit{J Intell Manuf} 2025. doi:\href{https://doi.org/10.1007/s10845-025-02691-5}{10.1007/s10845-025-02691-5}.


\bibitem{yedurkar2025}
Yedurkar DP, Schlech T, Sause MGR. A systematic review on smart and predictive maintenance in tool condition monitoring. \textit{IEEE Access} 2025;13:106246–86. doi:\href{https://doi.org/10.1109/ACCESS.2025.3579204}{10.1109/ACCESS.2025.3579204}.

\bibitem{hameed2023}
Hameed MS, Schwung A. Graph neural networks-based scheduler for production planning problems using reinforcement learning. \textit{J Manuf Syst} 2023;69:91–102. doi:\href{https://doi.org/10.1016/j.jmsy.2023.06.005}{10.1016/j.jmsy.2023.06.005}.

\bibitem{feng2025}
Feng L. Joint optimization algorithm for vehicle scheduling and supply chain inventory management based on multi-agent deep reinforcement learning. \textit{Neural Comput Appl} 2025. doi:\href{https://doi.org/10.1007/s00521-025-11661-0}{10.1007/s00521-025-11661-0}.

\bibitem{zeynivand2025}
Zeynivand M, Esmaili P, Cristaldi L, Gruosso G. A novel approach to digital twin-based energy efficiency monitoring and failure analysis in industrial applications. \textit{J Manuf Syst} 2025;83:612–25. doi:\href{https://doi.org/10.1016/j.jmsy.2025.10.011}{10.1016/j.jmsy.2025.10.011}.

\bibitem{wan2025}
Wan N, Wang Z, Li M, Liu Q, Zhao B, Ding W, Xu J. Design and optimization of grinding tool structures for high-performance machining of aerospace difficult-to-cut materials. \textit{J Manuf Process} 2025;154:640–58. doi:\href{https://doi.org/10.1016/j.jmapro.2025.10.006}{10.1016/j.jmapro.2025.10.006}.

\bibitem{luo2025}
Luo Q, Huang N, Bartles DL, et al. Geometry-informed multimodal variational autoencoder for real-time prediction of properties for Ti–6Al–4V fabricated using PBF-LB. \textit{J Intell Manuf} 2025. doi:\href{https://doi.org/10.1007/s10845-025-02693-3}{10.1007/s10845-025-02693-3}.

\bibitem{gultekin2025}
Gültekin Ö, Cinar E. Deep meta-learning-based multi-signal data fusion approach for fault diagnosis. \textit{J Intell Manuf} 2025. doi:\href{https://doi.org/10.1007/s10845-025-02609-1}{10.1007/s10845-025-02609-1}.

\bibitem{tanveer2025}
Tanveer M, Azad MM, Kim D, et al. Generative design for engineering applications: a state-of-the-art review. \textit{Arch Comput Methods Eng} 2025. doi:\href{https://doi.org/10.1007/s11831-025-10302-y}{10.1007/s11831-025-10302-y}.

\bibitem{wang2024}
Wang Z, Xu H. Manufacturability-aware deep generative design of 3D metamaterial units for additive manufacturing. \textit{Struct Multidisc Optim} 2024;67:22. doi:\href{https://doi.org/10.1007/s00158-023-03732-4}{10.1007/s00158-023-03732-4}.

\bibitem{cai2025}
Cai Q, Ma J, Xie YM, San B, Zhou Y. Topology optimization and diverse truss designs considering nodal stability and bar buckling. \textit{J Constr Steel Res} 2025;224(Pt A):109128. doi:\href{https://doi.org/10.1016/j.jcsr.2024.109128}{10.1016/j.jcsr.2024.109128}.

\bibitem{krahe2022similarity}
Krahe C, Marinov M, Schmutz T, Hermann Y, Bonny M, May M, Lanza G. AI based geometric similarity search supporting component reuse in engineering design. \textit{Procedia CIRP} 2022;109:275–280. doi:\href{https://doi.org/10.1016/j.procir.2022.05.249}{10.1016/j.procir.2022.05.249}.

\bibitem{wang2025}
Wang X, Lou H, Tang L, Zhang Q. Data-driven and decomposition-based multiobjective multitask optimization for automotive shape design problem. \textit{IEEE Trans Evol Comput} 2025. doi:\href{https://doi.org/10.1109/TEVC.2025.3580454}{10.1109/TEVC.2025.3580454}.

\bibitem{zhang2025cad}
Zhang C, Polette A, Pinquié R, Iida M, De Charnace H, Pernot J-P. Reinforcement learning-based parametric CAD models reconstruction from 2D orthographic drawings. \textit{Comput Aided Des} 2025;188:103925. doi:\href{https://doi.org/10.1016/j.cad.2025.103925}{10.1016/j.cad.2025.103925}.

\bibitem{jones2023cad}
Jones BT, Hu M, Kodnongbua M, Kim VG, Schulz A. Self-supervised representation learning for CAD. In: \textit{Proc IEEE/CVF Conf Comput Vis Pattern Recognit (CVPR)}. Vancouver, BC, Canada; 2023:21327–21336. doi:\href{https://doi.org/10.1109/CVPR52729.2023.02043}{10.1109/CVPR52729.2023.02043}.

\bibitem{li2025genai}
Li KY, Huang CK, Chen QW, et al. Generative AI and CAD automation for diverse and novel mechanical component designs under data constraints. \textit{Discov Appl Sci} 2025;7:315. doi:\href{https://doi.org/10.1007/s42452-025-06833-5}{10.1007/s42452-025-06833-5}.



\bibitem{xu2024}
Xu Z, Hong CS, Soria Zurita NF, Gyory JT, Stump G, Nolte H, Cagan J, McComb C. Adaptation through communication: Assessing human–artificial intelligence partnership for the design of complex engineering systems. \textit{J Mech Des} 2024;146(8):081401. doi:\href{https://doi.org/10.1115/1.4064490}{10.1115/1.4064490}.

\bibitem{picard2025}
Picard C, Edwards KM, Doris AC, et al. From concept to manufacturing: evaluating vision-language models for engineering design. \textit{Artif Intell Rev} 2025;58:288. doi:\href{https://doi.org/10.1007/s10462-025-11290-y}{10.1007/s10462-025-11290-y}.

\bibitem{heidari2025gdl}
Heidari N, Iosifidis A. Geometric deep learning for computer-aided design: a survey. \textit{IEEE Access} 2025;13:119305–119334. doi:\href{https://doi.org/10.1109/ACCESS.2025.3587121}{10.1109/ACCESS.2025.3587121}.

\bibitem{feng2019meshnet}
Feng Y, Feng Y, You H, Zhao X, Gao Y. MeshNet: Mesh neural network for 3D shape representation. \textit{Proc AAAI Conf Artif Intell} 2019;33(1):8279–8286. doi:\href{https://doi.org/10.1609/aaai.v33i01.33018279}{10.1609/aaai.v33i01.33018279}.

\bibitem{zhang2018featurenet}
Zhang Z, Jaiswal P, Rai R. FeatureNet: Machining feature recognition based on 3D convolution neural network. \textit{Comput Aided Des} 2018;101:12–22. doi:\href{https://doi.org/10.1016/j.cad.2018.03.006}{10.1016/j.cad.2018.03.006}.

\bibitem{le2018pointgrid}
Le T, Duan Y. PointGrid: A deep network for 3D shape understanding. \textit{Proc IEEE/CVF Conf Comput Vis Pattern Recognit (CVPR)} 2018;9204–9214. doi:\href{https://doi.org/10.1109/CVPR.2018.00959}{10.1109/CVPR.2018.00959}.

\bibitem{jayaraman2021uvnet}
Jayaraman PK, et al. UV-Net: Learning from boundary representations. \textit{Proc IEEE/CVF Conf Comput Vis Pattern Recognit (CVPR)} 2021;11698–11707. doi:\href{https://doi.org/10.1109/CVPR46437.2021.01153}{10.1109/CVPR46437.2021.01153}.

\bibitem{liu2024kan}
Liu Z, Wang Y, Vaidya S, Ruehle F, Halverson J, Soljačić M, Hou TY, Tegmark M. KAN: Kolmogorov–Arnold networks. \textit{arXiv preprint} arXiv:2404.19756, 2024. Available: \href{https://arxiv.org/abs/2404.19756}{2404.19756}.

\bibitem{pykan2024}
Z. Liu and M. Tegmark,
``Kolmogorov--Arnold Networks (PyKAN),'' 
Version~0.1.8, 2024. 
[Online]. Available: \href{https://github.com/KindXiaoming/pykan}{https://github.com/KindXiaoming/pykan}.


\bibitem{madan2007}
Madan J, Rao PVM, Kundra TK. Computer aided manufacturability analysis of die-cast parts. \textit{Comput Aided Des Appl} 2007;4:147–158. doi:\href{https://doi.org/10.1080/16864360.2007.10738535}{10.1080/16864360.2007.10738535}.

\bibitem{lynn2016}
Lynn R, Saldana C, Kurfess T, Reddy N, Simpson T, Jablokow K, Tucker T, Tedia S, Williams C. Toward rapid manufacturability analysis tools for engineering design education. \textit{Procedia Manuf} 2016;5:1183–1196. doi:\href{https://doi.org/10.1016/j.promfg.2016.08.093}{10.1016/j.promfg.2016.08.093}.



\bibitem{campana2020_1}
Campana G, Mele M. An application to stereolithography of a feature recognition algorithm for manufacturability evaluation. \textit{J Intell Manuf} 2020;31:199–214. doi:\href{https://doi.org/10.1007/s10845-018-1441-8}{10.1007/s10845-018-1441-8}.


\bibitem{winkler2021}
Winkler M, Stürmer S, Konrad C. Evaluation of the additive manufacturability of CAD-parts for initial data labelling in AI-based part identification. \textit{IOP Conf Ser Mater Sci Eng} 2021;1097(1):012020. doi:\href{https://doi.org/10.1088/1757-899X/1097/1/012020}{10.1088/1757-899X/1097/1/012020}.


\bibitem{xu2022}
Xu T, Xue J, Chen Z, et al. A systematic method for automated manufacturability analysis of machining parts. \textit{Int J Adv Manuf Technol} 2022;122:391–407. doi:\href{https://doi.org/10.1007/s00170-022-09586-2}{10.1007/s00170-022-09586-2}.


\bibitem{stavropoulos2022}
Stavropoulos P, Tzimanis K, Souflas T, et al. Knowledge-based manufacturability assessment for optimization of additive manufacturing processes based on automated feature recognition from CAD models. \textit{Int J Adv Manuf Technol} 2022;122:993–1007. doi:\href{https://doi.org/10.1007/s00170-022-09948-w}{10.1007/s00170-022-09948-w}.


\bibitem{korosec2005}
Korosec M, Balic J, Kopac J. Neural network based manufacturability evaluation of free form machining. \textit{Int J Mach Tools Manuf} 2005;45(1):13–20. doi:\href{https://doi.org/10.1016/j.ijmachtools.2004.06.022}{10.1016/j.ijmachtools.2004.06.022}.







\bibitem{ghadai2018_3DCNN}
Ghadai S, Balu A, Sarkar S, Krishnamurthy A. Learning localized features in 3D CAD models for manufacturability analysis of drilled holes. \textit{Comput Aided Geom Des} 2018;62:263–275. doi:\href{https://doi.org/10.1016/j.cagd.2018.03.024}{10.1016/j.cagd.2018.03.024}.



\bibitem{peddireddy2021_3DCNNs}
Peddireddy D, Fu X, Shankar A, Wang H, Joung BG, Aggarwal V, Sutherland JW, Jun MBG. Identifying manufacturability and machining processes using deep 3D convolutional networks. \textit{J Manuf Process} 2021;64:1336–1348. doi:\href{https://doi.org/10.1016/j.jmapro.2021.02.034}{10.1016/j.jmapro.2021.02.034}.



\bibitem{yan2022_3DVAEGANs}
Yan X, Melkote S. Generative modeling of the shape transformation capability of machining processes. \textit{Manuf Lett} 2022;33(Suppl):794--801. \href{https://doi.org/10.1016/j.mfglet.2022.07.098}{10.1016/j.mfglet.2022.07.098}.


\bibitem{yan2023_AESNN}
Yan X, Melkote S. Automated manufacturability analysis and machining process selection using deep generative model and Siamese neural networks. \textit{J Manuf Syst} 2023;67:57–67. doi:\href{https://doi.org/10.1016/j.jmsy.2023.01.006}{10.1016/j.jmsy.2023.01.006}.



\bibitem{zhong2025}
Zhong F, Wang Y, Wang P-S, Lu L, Zhao H. DeepMill: Neural accessibility learning for subtractive manufacturing. \textit{arXiv preprint} arXiv:2502.06093, 2025. doi:\href{https://doi.org/10.48550/arXiv.2502.06093}{10.48550/arXiv.2502.06093}.



\bibitem{shi2018}
Shi Y, Zhang Y, Baek S, De Backer W, Harik R. Manufacturability analysis for additive manufacturing using a novel feature recognition technique. \textit{Comput Aided Des Appl} 2018;15:941–52. doi:\href{https://doi.org/10.1080/16864360.2018.1462574}{10.1080/16864360.2018.1462574}.

\bibitem{mycroft2020}
Mycroft W, Katzman M, Tammas-Williams S, et al. A data-driven approach for predicting printability in metal additive manufacturing processes. \textit{J Intell Manuf} 2020;31:1769–81. doi:\href{https://doi.org/10.1007/s10845-020-01541-w}{10.1007/s10845-020-01541-w}.

\bibitem{guo2021}
Guo Y, Lu W F, Fuh J Y H. Semi-supervised deep learning based framework for assessing manufacturability of cellular structures in direct metal laser sintering process. \textit{J Intell Manuf} 2021;32(2):347–59. doi:\href{https://doi.org/10.1007/s10845-020-01575-0}{10.1007/s10845-020-01575-0}.

\bibitem{zhang2021}
Zhang Y, Yang S, Dong G, Zhao YF. Predictive manufacturability assessment system for laser powder bed fusion based on a hybrid machine learning model. \textit{Addit Manuf} 2021;41:101946. doi:\href{https://doi.org/10.1016/j.addma.2021.101946}{10.1016/j.addma.2021.101946}.

\bibitem{zhang2022}
Zhang Y, Zhao YF. Hybrid sparse convolutional neural networks for predicting manufacturability of visual defects of laser powder bed fusion processes. \textit{J Manuf Syst} 2022;62:835–45. doi:\href{https://doi.org/10.1016/j.jmsy.2021.07.002}{10.1016/j.jmsy.2021.07.002}.

\bibitem{zhang2022web}
Zhang Y, Zhao YF. A web-based automated manufacturability analyzer and recommender for additive manufacturing (MAR-AM) via a hybrid machine learning model. \textit{Expert Syst Appl} 2022;199:117189. doi:\href{https://doi.org/10.1016/j.eswa.2022.117189}{10.1016/j.eswa.2022.117189}.



\bibitem{zimmerling2019}
Zimmerling C, Trippe D, Fengler B, Kärger L. An approach for rapid prediction of textile draping results for variable composite component geometries using deep neural networks. \textit{AIP Conf Proc} 2019;2113(1):020007. doi:\href{https://doi.org/10.1063/1.5112512}{10.1063/1.5112512}.

\bibitem{hwang2024}
Hwang Y-M, Ho T-H, Huang Y-F, Chen C-M. Formability prediction using machine learning combined with process design for high-drawing-ratio aluminum alloy cups. \textit{Materials} 2024;17(16):3991. doi:\href{https://doi.org/10.3390/ma17163991}{10.3390/ma17163991}.

\bibitem{siegfried2024}
Siegfried R, Villamizar M, Devènes S, Rey R, Bacha A, Morère B, Zahno S, Odobez JM. Designing deep learning-based tools leveraging production data to support manufacturability analysis. \textit{SSRN Electron J} 2024. doi:\href{https://doi.org/10.2139/ssrn.4741197}{10.2139/ssrn.4741197}.







\bibitem{nelaturi2015}
Nelaturi S, Kim W, Kurtoglu T. Manufacturability feedback and model correction for additive manufacturing. \textit{J Manuf Sci Eng} 2015;137(2):021015. doi:\href{https://doi.org/10.1115/1.4029374}{10.1115/1.4029374}.

\bibitem{cai2018}
Cai N, Bendjebla S, Lavernhe S, Mehdi-Souzani C, Anwer N. Freeform machining feature recognition with manufacturability analysis. \textit{Procedia CIRP} 2018;72:1475–80. doi:\href{https://doi.org/10.1016/j.procir.2018.03.261}{10.1016/j.procir.2018.03.261}.

\bibitem{gupta2019}
Gupta MK, Swain AK, Jain PK. A novel approach to recognize interacting features for manufacturability evaluation of prismatic parts with orthogonal features. \textit{Int J Adv Manuf Technol} 2019;105:343–73. doi:\href{https://doi.org/10.1007/s00170-019-04073-7}{10.1007/s00170-019-04073-7}.

\bibitem{chen2020}
Chen L, Lau TY, Tang K. Manufacturability analysis and process planning for additive and subtractive hybrid manufacturing of quasi-rotational parts with columnar features. \textit{Comput Aided Des} 2020;118:102759. doi:\href{https://doi.org/10.1016/j.cad.2019.102759}{10.1016/j.cad.2019.102759}.

\bibitem{coatanea2021}
Coatanéa E, Nagarajan HPN, Panicker S, et al. Systematic manufacturability evaluation using dimensionless metrics and singular value decomposition: a case study for additive manufacturing. \textit{Int J Adv Manuf Technol} 2021;115:715–31. doi:\href{https://doi.org/10.1007/s00170-020-06158-0}{10.1007/s00170-020-06158-0}.

\bibitem{zhao2022}
Zhao C, Ma C, Zhang H, et al. Modeling manufacturing resources based on manufacturability features. \textit{Sci Rep} 2022;12:10775. doi:\href{https://doi.org/10.1038/s41598-022-15072-2}{10.1038/s41598-022-15072-2}.




\bibitem{sommers2024}
Sommers A, Rahimi S, McCall T, Wall E, Henslee A, Dalton L, Babin PD, Watson N, Sharma G, Parmar MD. A hybrid expert system for estimation of the manufacturability of a notional design. \textit{Appl Comput Intell Soft Comput} 2024;2024:4985090. doi:\href{https://doi.org/10.1155/2024/4985090}{10.1155/2024/4985090}.

\bibitem{deep2025}
Deep A, Miri Beidokhti M, Piili H. Preliminary manufacturability evaluation of complex geometrical parts based on layer thickness in the metal powder bed fusion process. \textit{Prog Addit Manuf} 2025;10:8941–61. doi:\href{https://doi.org/10.1007/s40964-025-01202-5}{10.1007/s40964-025-01202-5}.



\bibitem{Sun2017}
Sun C, Shrivastava A, Singh S, Gupta A. Revisiting unreasonable effectiveness of data in deep learning era. \textit{Proc IEEE Int Conf Comput Vis} 2017;843–52. doi:\href{https://doi.org/10.1109/ICCV.2017.97}{10.1109/ICCV.2017.97}.

\bibitem{bhatt2024}
Bhatt N, Bhatt N, Prajapati P, et al. A data-centric approach to improve performance of deep learning models. \textit{Sci Rep} 2024;14:22329. doi:\href{https://doi.org/10.1038/s41598-024-73643-x}{10.1038/s41598-024-73643-x}.


\bibitem{Recht2019}
Recht B, Roelofs R, Schmidt L, Shankar V. Do ImageNet classifiers generalize to ImageNet? \textit{Proc Int Conf Mach Learn} 2019;97:5389–5400. \href{https://proceedings.mlr.press/v97/recht19a.html}{proceedings.mlr.press/v97/recht19a.html}.

\bibitem{khinvasara2024}
Khinvasara T, Ness S, Shankar A. Leveraging AI for enhanced quality assurance in medical device manufacturing. \textit{Asian J Res Comput Sci} 2024;17(6):13–35. doi:\href{https://doi.org/10.9734/ajrcos/2024/v17i6454}{10.9734/ajrcos/2024/v17i6454}.

\bibitem{colombi2024}
Colombi L, et al. Optimizing Industry 5.0 machine learning-based applications via synthetic data generation. In: \textit{Proc IEEE CAMAD}. Athens, Greece; 2024. p.~1–6. doi:\href{https://doi.org/10.1109/CAMAD62243.2024.10942898}{10.1109/CAMAD62243.2024.10942898}.


\bibitem{zha2025}
Zha D, Bhat ZP, Lai K-H, Yang F, Jiang Z, Zhong S, Hu X. Data-centric artificial intelligence: a survey. \textit{ACM Comput Surv} 2025;57(5):129. doi:\href{https://doi.org/10.1145/3711118}{10.1145/3711118}.


\bibitem{zimmerling2022}
Zimmerling C, Fengler B, Kärger L. Formability assessment of variable geometries using machine learning—analysis of the influence of the database. \textit{Key Eng Mater} 2022;926:2247–2257. doi:\href{https://doi.org/10.4028/p-1o0007}{10.4028/p-1o0007}.


\bibitem{wu2015shapenets}
Wu Z, Song S, Khosla A, Yu F, Zhang L, Tang X, Xiao J. 3D ShapeNets: A deep representation for volumetric shapes. \textit{Proc IEEE Conf Comput Vis Pattern Recognit (CVPR)} 2015;1912–1920. doi:\href{https://doi.org/10.1109/CVPR.2015.7298801}{10.1109/CVPR.2015.7298801}.

\bibitem{Liu2019} 
Z. Liu, H. Tang, Y. Lin, and S. Han,  
``Point-Voxel CNN for Efficient 3D Deep Learning,''  
\textit{Advances in Neural Information Processing Systems (NeurIPS)},  
pp.~9638--9648, 2019.  
doi:\href{https://papers.neurips.cc/paper/8382-point-voxel-cnn-for-efficient-3d-deep-learning.pdf}{papers.neurips.cc/point-voxel-cnn}.

\bibitem{Ghadai2021} 
S. Ghadai, L. Zhang, J. Gao, and C. McMillin,  
``Multi-resolution 3D convolutional neural network for learning multi-scale spatial features,''  
\textit{Computer Aided Geometric Design},  
vol.~88, p.~102027, 2021.  
doi:\href{https://doi.org/10.1016/j.cagd.2021.102027}{10.1016/j.cagd.2021.102027}.

\bibitem{Grosseheide2023} 
J. Großeheide, S. Fahr, S. Hoffmann, and A. Hoffmann,  
``Voxel-based description model of quality-related data for a volumetric voxel model representing tolerances,''  
\textit{euspen -- The European Society for Precision Engineering and Nanotechnology},  
2023.  
Available: \href{https://www.euspen.eu/knowledge-base/AM23125.pdf}{euspen.eu/knowledge-base/AM23125.pdf}.

\bibitem{Hilbig2023} 
A. Hilbig, C. Fleißner, F. Fröhlich, M. Dübel, M. Lieb, and P. Sachs,  
``Enhancing three-dimensional convolutional neural network-based geometric reasoning for engineering parts: addressing discretization and resolution challenges,''  
\textit{Journal of Computational Design and Engineering},  
vol.~10, no.~3, pp.~992--1006, 2023.  
doi:\href{https://doi.org/10.1093/jcde/qwad033}{10.1093/jcde/qwad033}.

\bibitem{Fei2024} 
Y. Fei and Y. Li,  
``Rotation invariance and equivariance in 3D deep learning: A comprehensive review,''  
\textit{Artificial Intelligence Review},  
vol.~57, no.~6, pp.~1--42, 2024.  
doi:\href{https://doi.org/10.1007/s10462-024-10741-2}{10.1007/s10462-024-10741-2}.

\bibitem{Mumuni2022} 
A. Mumuni and M. Mumuni,  
``Data augmentation: A comprehensive survey of modern approaches,''  
\textit{Journal of Big Data},  
vol.~9, no.~1, pp.~1--55, 2022.  
doi:\href{https://doi.org/10.1186/s40537-022-00660-4}{10.1186/s40537-022-00660-4}.

\bibitem{Maturana2015} 
D. Maturana and S. Scherer,  
``VoxNet: A 3D Convolutional Neural Network for Real-Time Object Recognition,''  
\textit{Proceedings of the IEEE/RSJ International Conference on Intelligent Robots and Systems (IROS)},  
pp.~922--928, 2015.  
doi:\href{https://doi.org/10.1109/IROS.2015.7353481}{10.1109/IROS.2015.7353481}.

\bibitem{Wu2015} 
Z. Wu, S. Song, A. Khosla, F. Yu, L. Zhang, X. Tang, and J. Xiao,  
``3D ShapeNets: A deep representation for volumetric shapes,''  
\textit{Proceedings of the IEEE Conference on Computer Vision and Pattern Recognition (CVPR)},  
pp.~1912--1920, 2015.  
doi:\href{https://doi.org/10.1109/CVPR.2015.7298801}{10.1109/CVPR.2015.7298801}.





\bibitem{kalpakjian2021_ch24}
S. Kalpakjian and S. R. Schmid,
\textit{Manufacturing Engineering and Technology}, 7th~ed. (SI Edition), ``Chapter~24,'' pp.~668--683,
Pearson Education, 2021.
ISBN: 978-9810694067.


\bibitem{boothroyd2010_mill}
G. Boothroyd, P. Dewhurst, and W. A. Knight,
\textit{Product Design for Manufacture and Assembly}, 3rd~ed., ``Chapter~7: Design for Machining,'' p.~293--301,
CRC Press, Boca Raton, FL, 2010.
doi:\href{https://doi.org/10.1201/9781420089288}{10.1201/9781420089288}.
ISBN: 978-0-429-14296-3.

\bibitem{machinery2020_3}
E. Oberg, F. D. Jones, H. L. Horton, H. H. Ryffel, and C. J. McCauley,
\textit{Machinery’s Handbook}, 31st ed., ``Milling Cutters,'' pp.~871--898,
Industrial Press, Inc., New York, 2020.

\bibitem{kalpakjian2021_ch23}
S. Kalpakjian and S. R. Schmid,
\textit{Manufacturing Engineering and Technology}, 7th~ed. (SI Edition), ``Chapter~23,'' pp.~653--655,
Pearson Education, 2021.
ISBN: 978-9810694067.

\bibitem{msc2019}
MSC Industrial Supply Co. \textit{The Big Book Catalog}, Version 24. Melville, NY, USA; 2019. Available at: \href{https://dirxion.mscdirect.com/bigbook/2019?BookCode=bb919flx&Lang=enu&origin=www.mscdirect.com}{The Big Book Catalog}.


\bibitem{boothroyd2010_drill}
G. Boothroyd, P. Dewhurst, and W. A. Knight,
\textit{Product Design for Manufacture and Assembly}, 3rd~ed., ``Chapter~7: Design for Machining,'' p.~309,
CRC Press, Boca Raton, FL, 2010.
doi:\href{https://doi.org/10.1201/9781420089288}{10.1201/9781420089288}.
ISBN: 978-0-429-14296-3.

\bibitem{machinery2020_1}
E. Oberg, F. D. Jones, H. L. Horton, H. H. Ryffel, and C. J. McCauley,
\textit{Machinery’s Handbook}, 31st ed., ``Twist Drills,'' pp.~931--962,
Industrial Press, Inc., New York, 2020.

\bibitem{machinery2020_2}
E. Oberg, F. D. Jones, H. L. Horton, H. H. Ryffel, and C. J. McCauley,
\textit{Machinery’s Handbook}, 31st ed., ``MACHINING NONFERROUS METALS
AND NON-METALLIC MATERIALS,'' pp.~1257,
Industrial Press, Inc., New York, 2020.

\bibitem{asm1989_drill}
ASM International,
\textit{ASM Handbook, Volume 16: Machining}, ``Drilling,'' p.~441,
ASM International, Metals Park, OH, 1989.
ISBN: 978-0-87170-022-3.

\bibitem{bralla_ch3_7} 
J. G. Bralla (Ed.),
``Design for Manufacturability Handbook, 2nd ed., Chapter 3.7: pp.~3.70--3.72,''
McGraw-Hill, New York, 1998.
ISBN: 0-07-007139-X.

\bibitem{bralla_ch4_5} 
J. G. Bralla (Ed.),
``Design for Manufacturability Handbook, 2nd ed., Chapter 4.5: pp.~4.45--4.55,''
McGraw-Hill, New York, 1998.
ISBN: 0-07-007139-X.

\bibitem{bralla_ch6_10} 
J. G. Bralla (Ed.),
``Design for Manufacturability Handbook, 2nd ed., Chapter 6.10: pp.~6.157--6.158,''
McGraw-Hill, New York, 1998.
ISBN: 0-07-007139-X.

\bibitem{iscar2024}
ISCAR Ltd. \textit{Drilling Handbook – General Purpose Drilling}. ISCAR’s Reference Guide, p.~48. Available: \href{https://www.iscar.com/Catalogs/Publication/english_1/Drilling_Handbook/Drilling%20Handbook-for%20web.pdf}{ISCAR’s Reference Guide}.


\bibitem{sandvik_drilling}
Sandvik Coromant Academy,
\textit{Metal Cutting Technology Training Handbook}, ``Drilling,'' pp.~E4--E47,
Sandvik Coromant, 2020.


\bibitem{groover_ch22}
M. P. Groover,
``Fundamentals of Modern Manufacturing: Materials, Processes, and Systems, 7th ed., Chapter 22: pp.~521--523,''
Wiley, Hoboken, NJ, 2020.
ISBN: 978-1-119-72201-4.


\bibitem{iso2768_1_1989}
International Organization for Standardization (ISO),
\textit{ISO 2768-1:1989 – General Tolerances, Part 1: Tolerances for Linear and Angular Dimensions Without Individual Tolerance Indications},
1st~ed., p.~2,
ISO, Geneva, Switzerland, 1989.
[Confirmed 2022].


\bibitem{groover_ch23}
M. P. Groover,
``Fundamentals of Modern Manufacturing: Materials, Processes, and Systems, 7th ed., Chapter 23: pp.~543--544,''
Wiley, Hoboken, NJ, 2020.
ISBN: 978-1-119-72201-4.


\bibitem{bralla_ch1_4}
J. G. Bralla (Ed.),
``Design for Manufacturability Handbook, 2nd ed., Chapter 1.4: p.~1.29,''
McGraw-Hill, New York, 1998.
ISBN: 0-07-007139-X.

\bibitem{asm1989_mill}
ASM International,
\textit{ASM Handbook, Volume 16: Machining}, ``Milling,'' p.~697,
ASM International, Metals Park, OH, 1989.
ISBN: 978-0-87170-022-3.




\bibitem{kolmogorov1957representation}
A. N. Kolmogorov,
`` On the representation of continuous functions of many vari-ables by superposition of continuous functions of one variable and addition,''
\textit{Doklady Akademii Nauk SSSR},
vol.~114, no.~5, pp.~953--956, 1957.


\bibitem{kolmogorov1961representation}
V. I. Arnold,
``On the representation of continuous functions of several vari-ables by superpositions of continuous functions of a smaller number of variables,''
\textit{American Mathematical Society}, 1963.




\bibitem{rumelhart1986}
Rumelhart D, Hinton G, Williams R. Learning representations by back-propagating errors. \textit{Nature} 1986;323:533–36. \href{https://doi.org/10.1038/323533a0}{10.1038/323533a0}.

\bibitem{bishop1995}
Bishop CM. Neural networks for pattern recognition. Oxford: Oxford University Press; 1995. \href{https://doi.org/10.1093/oso/9780198538493.001.0001}{10.1093/oso/9780198538493.001.0001}.

\bibitem{chen2016xgboost}
Chen T, Guestrin C. XGBoost: a scalable tree boosting system. \textit{Proc ACM SIGKDD Int Conf Knowl Discov Data Min} 2016;785–94. \href{https://doi.org/10.1145/2939672.2939785}{10.1145/2939672.2939785}.

\bibitem{ke2017lightgbm}
Ke G, Meng Q, Finley T, Wang T, Chen W, Ma W, Ye Q, Liu T-Y. LightGBM: a highly efficient gradient boosting decision tree. \textit{Adv Neural Inf Process Syst} 2017;30. \href{https://proceedings.neurips.cc/paper_files/paper/2017/file/6449f44a102fde848669bdd9eb6b76fa-Paper.pdf}{proceedings.neurips.cc/LightGBM.pdf}.

\bibitem{friedman2001gbm}
Friedman JH. Greedy function approximation: a gradient boosting machine. \textit{Ann Stat} 2001;29(5):1189--1232. \href{https://doi.org/10.1214/aos/1013203451}{10.1214/aos/1013203451}.

\bibitem{prokhorenkova2018catboost}
Prokhorenkova L, Gusev G, Vorobev A, Dorogush AV, Gulin A. CatBoost: unbiased boosting with categorical features. \textit{Adv Neural Inf Process Syst} 2018;31. \href{https://proceedings.neurips.cc/paper_files/paper/2018/file/14491b756b3a51daac41c24863285549-Paper.pdf}{proceedings.neurips.cc/CatBoost.pdf}.

\bibitem{geurts2006extratrees}
Geurts P, Ernst D, Wehenkel L. Extremely randomized trees. \textit{Mach Learn} 2006;63:3–42. \href{https://doi.org/10.1007/s10994-006-6226-1}{10.1007/s10994-006-6226-1}.

\bibitem{breiman2001rf}
Breiman L. Random forests. \textit{Mach Learn} 2001;45:5–32. \href{https://doi.org/10.1023/A:1010933404324}{10.1023/A:1010933404324}.

\bibitem{breiman1984cart}
Breiman L, Friedman J, Olshen RA, Stone CJ. Classification and regression trees. 1st ed. Chapman and Hall/CRC; 1984. \href{https://doi.org/10.1201/9781315139470}{10.1201/9781315139470}.


\bibitem{cover1967nn}
Cover T, Hart P. Nearest neighbor pattern classification. \textit{IEEE Trans Inf Theory} 1967;13(1):21–7. \href{https://doi.org/10.1109/TIT.1967.1053964}{10.1109/TIT.1967.1053964}.

\bibitem{hoerl1970ridge}
Hoerl AE, Kennard RW. Ridge regression: biased estimation for nonorthogonal problems. \textit{Technometrics} 1970;12(1):55–67. \href{https://doi.org/10.1080/00401706.1970.10488634}{10.1080/00401706.1970.10488634}.

\bibitem{cox1958}
Cox DR. The regression analysis of binary sequences. \textit{J R Stat Soc Ser B Methodol} 1958;20(2):215–42. \href{http://www.jstor.org/stable/2983890}{http://www.jstor.org/stable/2983890}.

\bibitem{cortes1995}
Cortes C, Vapnik V. Support-vector networks. \textit{Mach Learn} 1995;20:273–97. \href{https://doi.org/10.1007/BF00994018}{10.1007/BF00994018}.

\bibitem{hand2001}
Hand DJ, Yu K. Idiot’s Bayes: not so stupid after all? \textit{Int Stat Rev} 2001;69(3):385–98. \href{https://doi.org/10.2307/1403452}{10.2307/1403452}.



\end{thebibliography}
\end{document}